\documentclass[preprint,12pt,authoryear]{elsarticle}

\usepackage{amssymb}
\usepackage{amsmath}
\usepackage{amsthm}

\usepackage{algorithmic}
\usepackage{algorithm}
\usepackage{bm}
\usepackage{array}
\usepackage{subfig}
\usepackage{textcomp}
\usepackage{stfloats}
\usepackage{url}
\usepackage{verbatim}
\usepackage{graphicx}
\usepackage{booktabs}
\usepackage[table]{xcolor}
\usepackage{float}
\usepackage{makecell}

\usepackage{wrapfig}

\usepackage{multirow}

\usepackage{array}
\newcolumntype{L}[1]{>{\raggedright\let\newline\\\arraybackslash\hspace{0pt}}m{#1}}
\newcolumntype{C}[1]{>{\centering\let\newline\\\arraybackslash\hspace{0pt}}m{#1}}
\newcolumntype{R}[1]{>{\raggedleft\let\newline\\\arraybackslash\hspace{0pt}}m{#1}}

\journal{Neural Networks}

\begin{document}

\begin{frontmatter}



\title{Forward-Forward Learning achieves Highly Selective Latent Representations for Out-of-Distribution Detection in Fully Spiking Neural Networks} 

\author[tecnalia,deusto]{Erik B. Terres-Escudero} 
\ead{erik.terres@tecnalia.com}
\author[tecnalia,upv]{Javier Del Ser}
\author[ideko]{Aitor Martinez-Seras}
\author[deusto]{Pablo García Bringas}
\affiliation[tecnalia]{organization={TECNALIA},
            city={Derio},
            postcode={48160}, 
            country={Spain}}
\affiliation[deusto]{organization={Faculty of Engineering, University of Deusto},
            city={Bilbao},
            postcode={48007}, 
            country={Spain}}
\affiliation[upv]{organization={Dept. of Mathematics, University of the Basque Country (UPV/EHU)},
            city={Leioa},
            postcode={48940}, 
            country={Spain}}
\affiliation[ideko]{organization={Ideko S.Coop},
            city={Elgoibar},
            postcode={20870}, 
            country={Spain}}

\begin{abstract}
    In recent years, Artificial Intelligence (AI) models have achieved remarkable success across various domains, yet challenges persist in two critical areas: ensuring robustness against uncertain inputs and drastically increasing model efficiency during training and inference. Spiking Neural Networks (SNNs), inspired by biological systems, offer a promising avenue for overcoming these limitations.  By operating in an event-driven manner, SNNs achieve low energy consumption and can naturally implement biological methods known for their high noise tolerance. In this work, we explore the potential of the spiking Forward-Forward Algorithm (FFA) to address these challenges, leveraging its representational properties for both Out-of-Distribution (OoD) detection and interpretability. To achieve this, we exploit the sparse and highly specialized neural latent space of FF networks to estimate the likelihood of a sample belonging to the training distribution. Additionally, we propose a novel, gradient-free attribution method to detect features that drive a sample away from class distributions, addressing the challenges posed by the lack of gradients in most visual interpretability methods for spiking models. We evaluate our OoD detection algorithm on well-known image datasets (e.g., Omniglot, Not-MNIST, CIFAR10), outperforming previous methods proposed in the recent literature for OoD detection in spiking networks. Furthermore, our attribution method precisely identifies salient OoD features, such as artifacts or missing regions, hence providing a visual explanatory interface for the user to understand why unknown inputs are identified as such by the proposed method.
\end{abstract}

\begin{graphicalabstract}
\includegraphics[width=\linewidth]{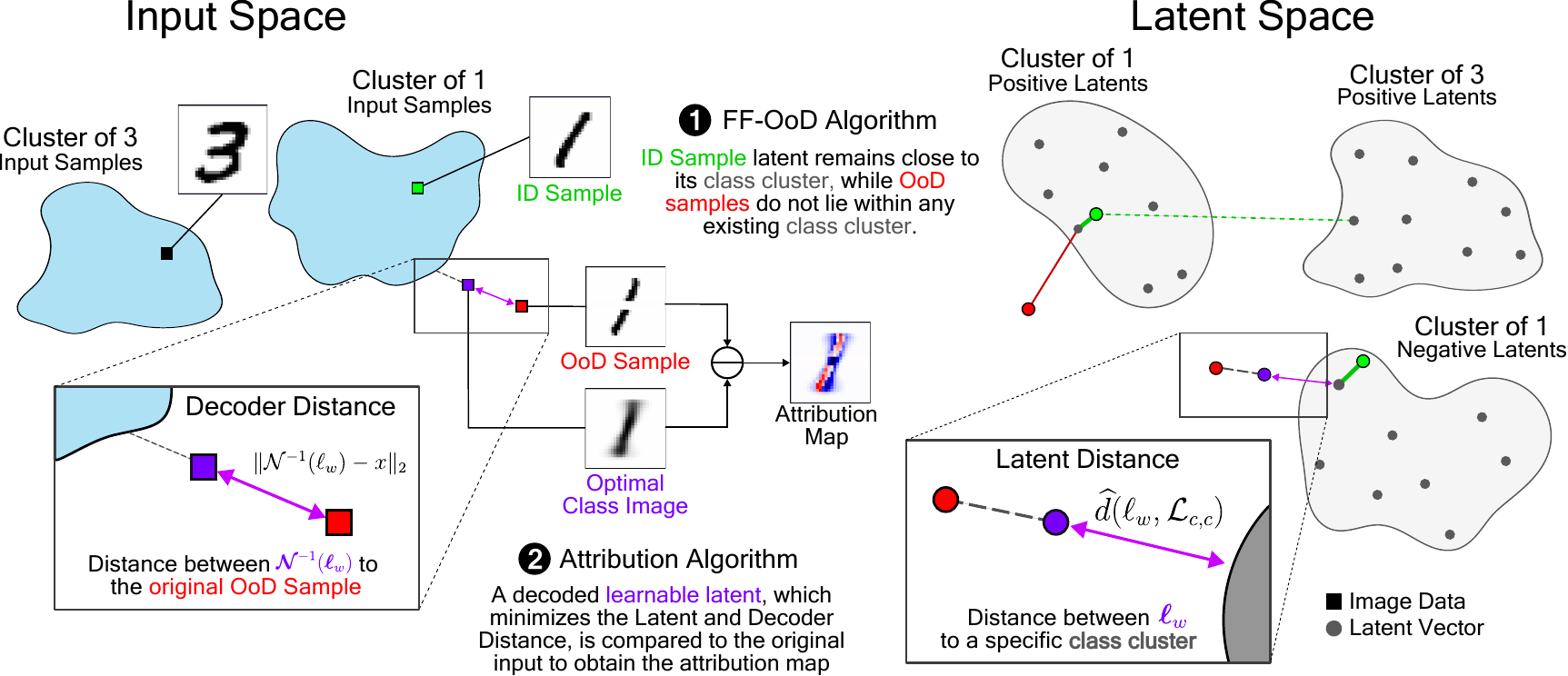}
\end{graphicalabstract}

\begin{highlights}
    \item Surrogate gradients allow for spiking adaptations of the Forward-Forward Algorithm
    \item The topology of Forward-Forward networks splits in and out of distribution samples
    \item Highly selective neurons can produce highly interpretable attribution maps
\end{highlights}

\begin{keyword}
Spiking Neural Networks\sep  Out-of-Distribution Detection \sep  Forward-Forward Algorithm \sep Explainable Artificial Intelligence



\end{keyword}

\end{frontmatter}


\section{Introduction}
\label{sec1}
The fast development of learning algorithms has drastically impacted our dependency on Artificial Intelligence (AI) models, whose exceptional capabilities have led to their continual integration into production-ready software. However, this recent surge has been heavily hindered by two main factors: first, the vast computational power required to train these models; and second, the lack of robustness guarantees to ensure that these models function reliably on novel or unexpected data. In contrast to these systems, biological brains operate using a negligible amount of energy, while demonstrating a strong capacity to learn and adapt to different data streams \citep{zador2022toward}. Recent research efforts have focused on developing highly parallel event-driven systems, inspired by the spike-driven dynamics of biological neurons, with the goal of providing an energy-efficient alternative to traditional GPUs \citep{tavanaei2019deep}. Due to their architectural design, these computers, denoted as neuromorphic computers, operate as a natural physical implementation of Spiking Neural Networks (SNNs), thereby obtaining higher inference speeds. In addition, due to their natural resemblance to biological brains, SNNs facilitate the use and development of biologically-inspired algorithms, which are known to exhibit a natural resistance to Out-of-Distribution (OoD) samples, providing an alternative mechanism to prevent undesired inputs and behavior in networks.
One recent algorithm following  these biological motivations is the so-called Forward-Forward Algorithm (FFA), which employs a local contrastive-based learning rule to make networks highly selective to In-Distribution (ID) data, while causing OoD samples to produce almost dormant latent activity \citep{hinton2022forward}. This learning heuristic, beyond demonstrating competitive accuracy with respect to backpropagation in dense networks, has also been shown to reproduce the highly representative properties of the latent space of biological models, which segregates the data into feature-selective clusters \citep{tosato2023emergent,ororbia2023predictive}. However, although some papers have proposed alternative implementations of this algorithm for SNNs \citep{ororbia2023learning,terres2024emerging}, the vast majority of research is still focused on the non-spiking original implementation. Similarly, the field of OoD detection is predominantly centered around algorithms developed for analog networks, which are not able to obtain the same quantitative results when adapted for SNNs \citep{seras2022novel}. In this context, the biological foundation of FFA, which makes it inherently suitable for SNNs, combined with its contrastive learning process that encourages OoD samples to diverge from ID inputs, presents a compelling synergy.

This work relies on a spiking implementation of the FFA to develop a fully-spiking OoD detection algorithm that exploits the geometrical properties of the latent space to define a distance-based scoring function. By providing empirical evidence of the large separation between ID and OoD clusters in the latent space, we propose a heuristic to determine whether an input belongs to the ID latent set. Given the importance of explainability mechanisms in the development of AI based systems \citep{diaz2023connecting}, we extend our results by proposing an algorithm that generates an attribution map for a given input, highlighting features that do not align with those expected from the ID samples of any desired class. The results presented in this paper demonstrate that spiking neural networks can achieve competitive robustness against unknown stimuli when integrated with biologically plausible frameworks, positioning them as a energy-efficient and safe modeling approach for responsible and trustworthy AI systems.

The rest of this paper is structured as follows: Section \ref{sec:relwork} provides a comprehensive review of related work. Section \ref{sec:methods} elaborates on all the methodological approaches developed in this work. Specifically, Section \ref{sec:approach_spiking_ffa} presents the approach taken to adapt FFA for SNNs, Section \ref{sec:approach_ff_scp} describes the FF-OoD algorithm, and Section \ref{sec:approach_attribution_technique} introduces our attribution algorithm. Next, Section \ref{sec:exp_setup} presents the research questions that this work addresses, along with the experimental setup used for the experiments. Subsequently, Section \ref{sec:res} presents the experimental results, along with a discussion on the observed limitations. Finally, Section \ref{sec:conc} outlines the conclusions of the paper and delineates directions for future research.

\section{Related Work and Contributions}\label{sec:relwork}

In this section, we present a comprehensive overview of relevant literature for this work, focusing on the following central topics:  Spiking Neural Networks (Section \ref{sec:related_work_spiking}); Forward-only Learning Algorithms (Section \ref{sec:rel_work_forward_only}); Out-of-Distribution Detection (Section \ref{sec:rel_work_ood}); and Explainability and Attribution (Section \ref{sec:related_work_attr}). Finally, Section \ref{sec:contrib} details the contribution that this papers offer with respect to the current literature.

\subsection{Spiking Neural Networks}\label{sec:related_work_spiking}

SNNs constitute a special category of artificial neural networks inspired by biological models, aiming to replicate the communication patterns found in real neurons within the brain. Unlike traditional approaches, primarily driven by Analog Neural Networks (ANNs) that rely on continuous activation values to forward information, spiking models operate by using discrete pulses of information, known as \emph{spikes}. The internal dynamics of the neuron, determined by the chosen neural model, dictate when spikes occur and are typically presented as a dynamical system that provides a mathematical approximation of the behavior observed in real neurons \citep{yamazaki2022spiking}. As such, depending on the degree of biological plausibility desired, models can be more accurate but computationally expensive (e.g., Hodgkin-Huxley), or they can satisfy multiple properties observed in real neurons while maintaining low computational complexity (e.g., Integrate-and-Fire models).

In this work, we employ the Leaky Integrate-and-Fire (LIF) neural model, which is known for preserving the standard action potential dynamics and being easily computable \citep{abbott1999lapicque,gerstner2014neuronal}. Under this model, the membrane potential \( U \) of the neuron is tracked and updated in response to the external stimulus \( I_{\text{in}} \). This activity decays over time by a decay factor $\tau_{t}$, creating the leaky dynamic that gives the model its name. The dynamics of the neuron can be described as follows:
\begin{equation}
    U(t) = \tau_{t} \cdot U(t-1) + R \cdot I_{\text{in}}(t),
\end{equation}
where \( U(t-1) \) represents the membrane potential at the previous time step, and \( R \) is a scaling factor of the input.

Once the membrane potential reaches a threshold \( \theta_{\text{thr}} \), a spike is generated and forwarded to subsequent neurons. At this point, the membrane potential is reset to zero, resulting in the following equations:
\begin{align}
    S(t) &= \mathbb{I}(U(t) > \theta_{\text{thr}}),\\
    U(t) &= U(t) \cdot (1 - S(t)),
\end{align}
where $\mathbb{I}(\cdot)$ is an auxiliary binary function taking value $1$ if its argument is true ($0$ otherwise).

\subsection{Forward-only Learning Algorithms} \label{sec:rel_work_forward_only}

Backpropagation stands as the most common algorithm for training artificial neural networks due to its high adaptability and precision on different architectures and loss objectives. However, the need for differentiable activations when creating the network poses a clear limitation in spiking models, which use non-differentiable neural models. To address this, the most commonly employed technique relies on surrogate gradients, which involve replacing the original neural model with a differentiable approximation during the backward pass, so that the gradient value remains close to the expected gradient direction \citep{neftci2019surrogate}. 

Recently, the Forward-Forward Algorithm was introduced as a forward-only alternative to backpropagation, advocating for a biologically-motivated heuristic in which the backward pass is replaced by an additional contrastive forward pass \citep{hinton2022forward}. In this second pass, the model is presented with an adversarial sample, which is used to contrastively learn to distinguish between samples from the training set (which we denote by \textit{positive} or $D_{\oplus}$) and synthetically created data (resp. \textit{negative} or $D_{\ominus}$). As it follows from its forward-only nature, each layer is trained independently, using a layer-specific loss that solely employs local information. To formulate these local updates, Hinton proposed the use of a \textit{goodness score}, which offers a numeric measurement related to the likelihood of a latent vector  belonging to the positive data distribution. This goodness function \( G : \mathbb{L} \rightarrow \mathbb{R} \) was defined as the squared Euclidean norm of the latent vector, where \(\mathbb{L}\) denotes the latent space of the given layer. Under this formulation, the goodness score serves as a measure of the layer's activity, thereby expecting highly active latent vectors when presented with positive samples, and nearly dormant vectors when encountering a negative instance. By combining this goodness score with a probability function \( P : \mathbb{R} \rightarrow [0,1] \), mapping the goodness domain into a probability range, the network can be trained by using the standard Binary Cross Entropy between positive and negative binary states. In the original work, this function was computed using a logistic function, modulated by a threshold value \( \theta \) to shift the scores closer to zero when presented with near-zero goodness scores. This probability is formally denoted as \( P(x; \theta) = \text{sigmoid}(x - \theta) \).

As previously stated, each learning step involves two contrastive forward passes, a positive pass and a negative pass. In its supervised formulation, positive instances are created by embedding the corresponding label of the input as a one-hot encoded vector into the corner of the base input image. Analogously, negative data is created by taking samples from the training dataset and embedding random labels that do not align with their corresponding class. Using this positive and negative generation method, the inference process computes the goodness score of the samples with respect to all labels, selecting the label with the highest goodness score as the predicted class. To enhance this process, \citet{lee2023symba} refined the label embedding process by associating each class with a random binary vector, which is then appended to each input, removing the overlap between the embedded labels and the data, achieving enhanced classification accuracy.

Despite its clear biological motivation, research on FFA is still heavily conducted in ANNs, with SNNs being explored in only a handful of papers. The first spiking adaptation was developed by \citet{ororbia2023learning}, who proposed replacing the learning rule with a Hebbian or STDP update through time, modulated by a positive/negative reward signal. Following this work, \citet{merkel2024contrastive} provided experimental evidence that Ororbia's biological adaptation yields positive results when training networks on physical neuromorphic chips. In a parallel effort, \citet{terres2024emerging} studied the relationship between FFA's formulation and neo-Hebbian frameworks, demonstrating how FFA represents a specific branch of contrastive neo-Hebbian learning rules. Nevertheless, research conducted with ANNs has highlighted ways to improve the method's accuracy and revealed the geometric structure that emerges in trained networks \citep{tosato2023emergent,ororbia2023predictive}. Studies from \citet{tosato2023emergent} or \citet{yang2023theory} consistently emphasize the characteristic latent topology produced by this learning heuristic, characterized by sparse latent activity and high neural specialization, resulting in a landscape where class clusters are distinctly separated by clear boundaries.

\subsection{Out-of-Distribution Detection} \label{sec:rel_work_ood}
OoD Detection plays a crucial role in assessing the security of production-ready AI pipelines, as it ensures that models aren't exposed to inputs that could produce unpredictable (and potentially dangerous) behavior. These algorithms operate by defining a scoring function that measures the likelihood of a sample from belonging to the ID dataset, which is then used as a filter to remove the examples that do not satisfy a user-defined likelihood threshold. The construction of these functions has been explored by using two different properties of the networks: their output probability distribution, and their internal latent space. Given the absence of a classification layer on FFA, this paper focuses on developing distance-based techniques, which provide with a distance function to the latent space to measure how far samples OoD lie with respect to the training distribution \citep{yang2021generalized}. Similarly, considering the lack of research on OoD Detection in forward-only methods, we offer a comparative perspective by choosing methods trained using backpropagation.

Among OoD Detection benchmarks, ODIN stands as the most widely-used classification-based method due to its accuracy and reduced computational costs. This algorithm operates by comparing the final classification distribution when scaling the softmax temperature and introducing small, gradient-based input perturbations, thereby making OoD samples produce lower confidence scores \citep{liang2017enhancing}. Within distance-based methods, \citet{bergman2020deep} proposed the deep k-nearest neighbor algorithm to measure the distance between input latent vectors and previously computed ID representatives, using these distances to detect anomalies. Building upon this work, \citet{sun2022out} extended the application of deep k-nearest neighbor distances for OoD detection, highlighting the importance of leveraging geometric properties in the latent space of trained networks for OoD detection tasks.

It is important to acknowledge that the OoD field has only recently been introduced into SNNs, with  \citet{seras2022novel} offering the first fully spiking OoD detection algorithm. In their work, they introduce the Spike Count Pattern (SPC) algorithm, which characterizes spike patterns across layers to identify whether a given input aligns with the latent behavior observed with ID data. In order to produce a fair comparison between works, and given the absence of other spiking OoD papers, we rely on an equivalent experimental setup as them.

\subsection{Interpretability} \label{sec:related_work_attr}

It is well acknowledged that neural networks operate as black boxes, achieving high performance across multiple tasks while offering little to no insight into the heuristics followed to obtain their outputs. The term \textit{Explainable AI} was recently coined to group all techniques focused on developing methods aimed at offering humanly understandable explanations for these inner mechanisms \citep{arrieta2020explainable}. Among the methods in this domain, \textit{Interpretability} techniques aim at providing heuristics for interpreting the inner behavior of a model. Within visual data, \textit{Grad-CAM}-like methods stands as one of the most impactful techniques, using gradient information extracted of the last convolutional layer to detect areas of the input with the highest impact on the prediction of a specific class \citep{selvaraju2017grad}. Nevertheless, due to its reliance on gradient information to obtain these attribution maps, such techniques remain elusive for spiking systems, where only a handful of methods have been proposed \citep{kim2021visual,bitar2023gradient}.

In their work, \citet{seras2022novel} present a technique suitable for spiking systems, involving a backward pass of a latent vector through the network to compute the relevance of specific features in the input space. This method operates by reconstructing spike patterns across the layers and enabling the quantification of feature activity, thereby identifying regions of significance within the image. The authors highlighted the importance of exploiting the representational capabilities of the latent space of non-convolutional networks to create interpretable techniques.

\subsection{Contributions} \label{sec:contrib}

This work introduces the FF-OoD algorithm, a spiking OoD detection method that leverages the characteristic latent topology observed in networks trained with FFA. While most existing literature on FFA has focused on its predictive capabilities in analog settings, this work takes an initial step toward exploring its applicability, paving the way for neuromorphic systems. Additionally, we propose an attribution technique that generates interpretable images, highlighting the features that push an image outside the distribution of any selected class. Through these developments, we aim to bridge the gap in the literature on OoD detection and interpretability in SNNs.

\section{Proposed Approach}\label{sec:methods}

This section introduces the methodological details of the algorithms proposed in this work. Given the differences from current approaches, Section \ref{sec:approach_spiking_ffa} outlines the adaptations employed in FFA to make it suitable for SNNs. Subsequently, Section \ref{sec:approach_ff_scp} introduces the FF-OoD algorithm, which relies on the latent geometry of networks trained with FFA to create an OoD detection algorithm. Lastly, Section \ref{sec:approach_attribution_technique} presents an attribution technique aimed at providing visual insights into the information linking arbitrary samples to the distribution of specific classes.

\subsection{Spiking Forward-Forward Algorithm}
\label{sec:approach_spiking_ffa}

\begin{figure*}[!t]
    \centering
    \includegraphics[width = \textwidth]{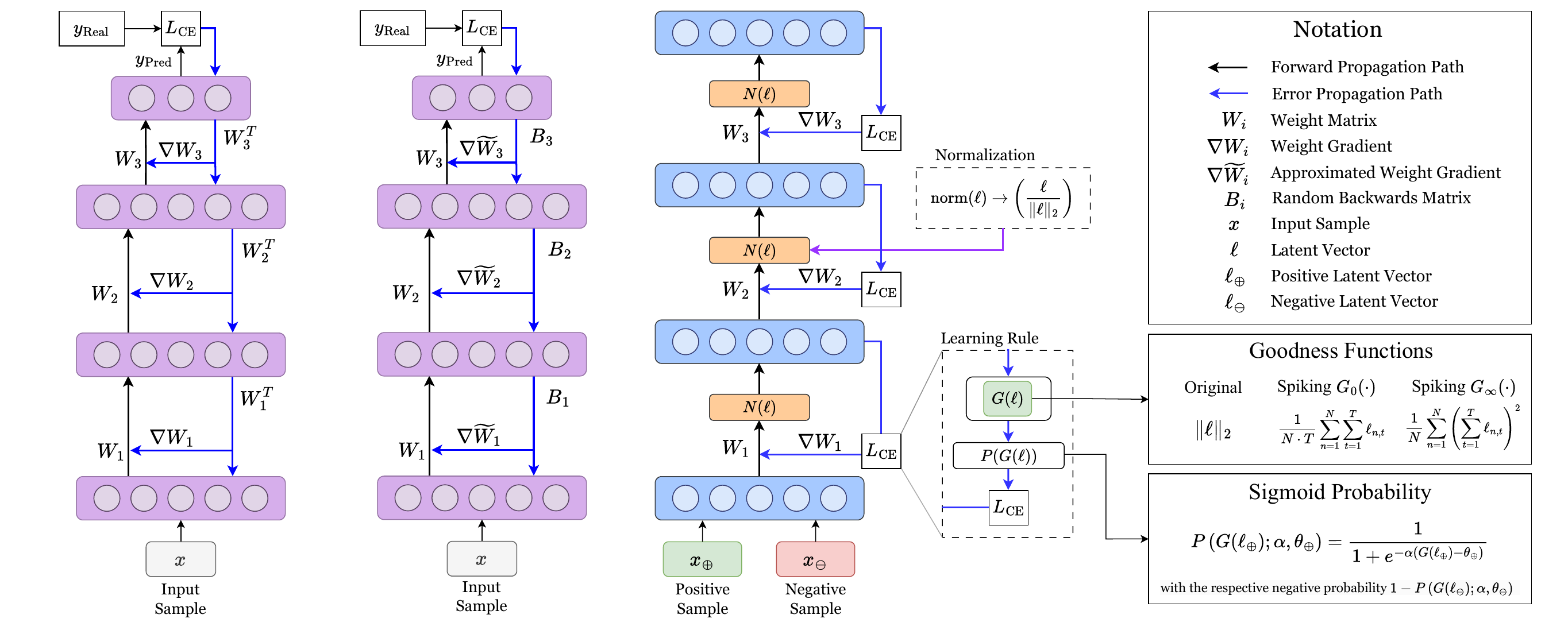}
    \caption{Diagram illustrating the architecture, from left to right, of Backpropagation (BP),  Feedback Alignment \citep{lillicrap2016random}, and Forward-Forward Algorithm (FFA) \citep{hinton2022forward}. Each architecture highlights the input forward path (black arrows), and the error propagation path (blue arrows). Additionally, we describe the update mechanism of FFA and its spiking adaptations.}
    \label{fig:ff_model_description}
\end{figure*}

Despite its original design as a biologically plausible alternative to Backpropagation, FFA has been mainly employed to train traditional ANNs. As previously mentioned, efforts have already been made to develop more biologically accurate alternatives, replacing its original error-based formula with a Hebbian Learning Rule. However, to avoid additional complexity within the network architecture, we advocate for the use of a surrogate gradient for the local gradient computation, which allows SNNs to be trained using the original learning heuristic. Following this design choice, the rest of the subsection details the modifications made to the goodness functions to adapt them to the new time dimension.

As detailed in Section \ref{sec:relwork}, the original goodness function \( G \) was formulated using the squared Euclidean distance of the latent vector. This function, combined with the unbounded dynamics of ReLU activations, enabled learning dynamics that could result in positive latent vectors achieving arbitrarily high goodness scores as training progressed. However, due to the discrete event-driven nature of SNNs, where the latent space consists of binary latent vectors, the activation value of each neuron is bounded by the total number of achievable spikes during the simulation period. Therefore, to replicate behaviors similar to those observed in ANNs, the goodness function must be adapted to handle these discrete states. We introduce two goodness functions for spiking models: an \textit{unbounded goodness}, which mimics the behavior of ANNs through the aggregation of the temporal dimension; and a \textit{bounded goodness}, designed to capitalize on the bounded activation range of SNNs.

\paragraph{Unbounded Goodness} This goodness function, referred to as \( G_{\infty}(\cdot) \), is obtained from the conventional goodness function by aggregating the spikes, resulting in an approximation of a ReLU-like function. In this scenario, the goodness function takes the latent vector \(\bm{\ell}\), composed of the spike train of the neurons of the layer, and acts as the squared Euclidean norm of the number of spikes. Formally, given the latent vector \(\bm{\ell} \in \{0,1\}^{N \times T}\) consisting of \(N\) neurons and \(T\) timesteps, the function \( G_{\infty}(\ell) : \{0, 1\}^{N \times T} \rightarrow \mathbb{R}_{\geq 0} \) is formulated as:
\begin{equation}
G_{\infty}(\bm{\ell}) = \frac{1}{N} \sum_{n=1}^{N} \left(\sum_{t=1}^{T} \ell_{n,t} \right)^2,
\end{equation}
where \(\ell_{n,t}\) denotes the value of the \(n\)-th neuron at timestep \(t\).

Although this function is not strictly unbounded, as it spans the interval \([0, N \cdot T^2]\), the upper limit of this interval can be reached by setting sufficiently high values of $N$ and $T$ to mimic the behavior of unbounded models.


\paragraph{Bounded Goodness} Our second goodness function, namely the \textit{bounded goodness}, denoted by \( G_{0}(\cdot) \), employs the average spiking rates of the neurons within the layer. By using this aggregation method, the obtained score remains independent of the number of timesteps of the simulation. Using the previous notation, the function \( G_{0}(\bm{\ell}) : \{0, 1\}^{N \times T} \rightarrow \mathbb{R}_{\geq 0} \) is expressed as:
\begin{equation}
G_{0}(\bm{\ell}) = \frac{1}{N \cdot T}\sum_{n=1}^{N}\sum_{t=1}^{T} \ell_{n,t}.
\end{equation}

Naturally, the bounded design of \( G_{0}(x) \) operates in a range limited to \([0, 1]\), regardless of the hyperparameter configuration. This bounded limit can pose a challenge in achieving extreme values for the probability function \( P(\cdot) \), especially if the original threshold \(\theta\) was intended for unbounded goodness scores. To address this, we modify the probability function in two ways. First, we introduce a scalar factor \( \alpha \) to scale the term \( G(\ell) - \theta \). Second, we replace the single threshold \( \theta \) with two thresholds, \( \theta_{\oplus} \) and \( \theta_{\ominus} \), applying independent thresholds to the positive and negative probability functions, respectively. This adjustment allows us to increase \( \alpha \) and adjust \( \theta_{\oplus}, \theta_{\ominus} \) to align with the domain of the probability function. Under this formulation, the resulting probability function is expressed as follows:
\begin{equation}
    \label{eq:probability_new}
    P(G(\ell_{\oplus}); \alpha, \theta_{\oplus}) = \frac{1}{1+e^{-\alpha  (G(\ell_{\oplus}) - \theta_{\oplus})}},
\end{equation}
where the converse negative probability function is defined by the expression $1-P(G(\ell_{\ominus}); \alpha, \theta_{\ominus})$.

\subsection{FF-OoD for Out-of-Distribution Detection}
\label{sec:approach_ff_scp}

\begin{figure*}[!t]
    \centering
    \includegraphics[width=\linewidth]{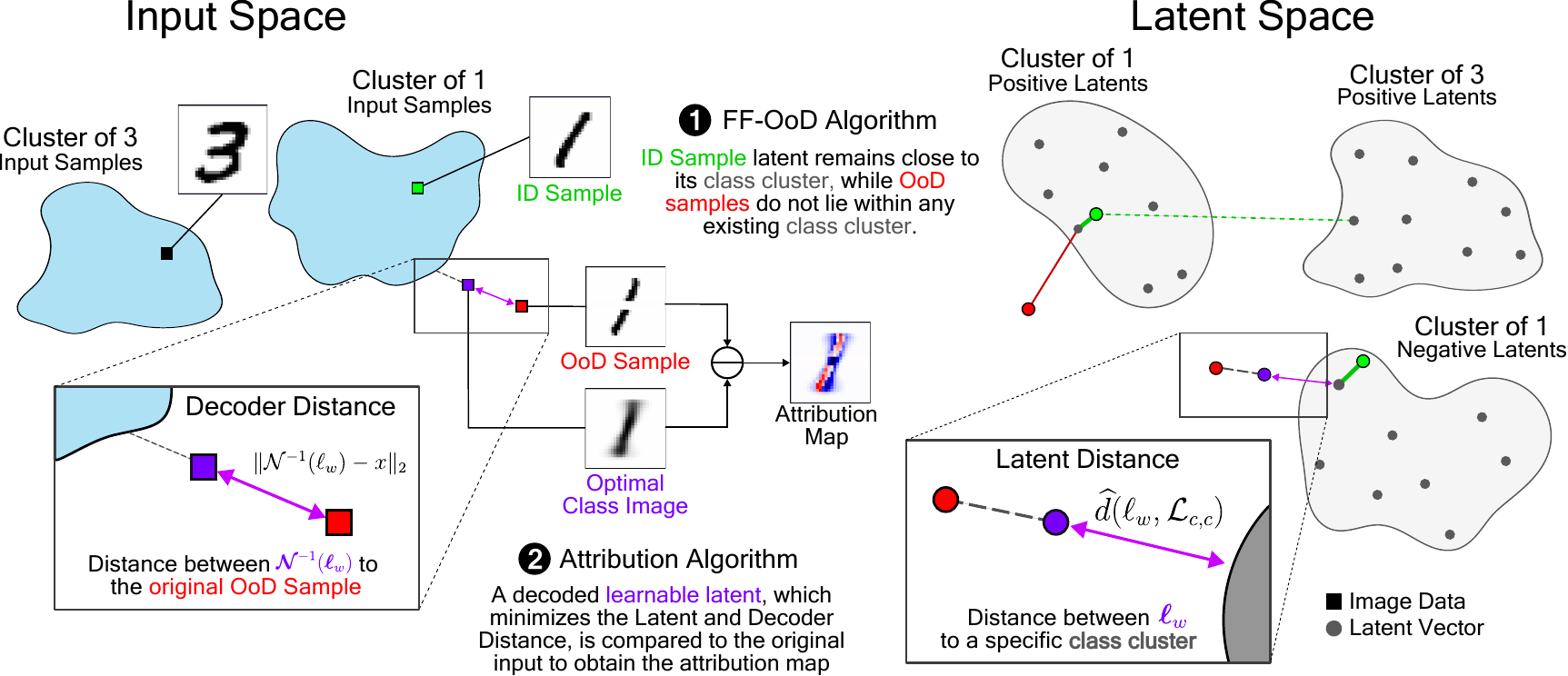}
    \caption{Visual diagram providing an example of a) the mechanics FF-OoD on an ID sample (green) and a OoD sample (red), with respect to the set of data from classes $1$ and $3$, and 2) the process of optimizing a latent vector (purple) so that when decoded, it can be compared with the original sample (the OoD sample) to obtain the atribution map to some specified class (Class 1).}
    \label{fig:attrb}
\end{figure*}

As shown in multiple works \citep{ororbia2023predictive,tosato2023emergent}, one key feature observed in networks trained using FFA is its emerging latent topology. This space has been characterized by a set of highly separated latent clusters, each related to a different class. This property, together with the observed sparsity of the latent vectors, showcases the discriminative capacity of the neurons within each layer of the network. In this section, we exploit this geometrical landscape in order to develop a OoD scoring function.

An additional property of FFA observed in our work is that class clustering is not solely confined to the positive latent space, but also appears in networks trained with bounded activations. Due to its original formulation, the negative goodness minimization process combined with ReLU activations resulted in the negative latent space being compressed into a small cluster around the zero vector. However, in scenarios where the inputs are noisy or when activations are bounded, the negative data exhibit non-zero latent vectors, which form their own clusters. These clusters can be characterized by the pair consisting of the real class label and the negative label embedded within the input. A visual representation of the emergent latent spaces is provided in \ref{ap:latent_geometry}.

Building on the geometrical motivation of SCP, we develop a distance-based algorithm that utilizes networks trained with FFA. In their work, Martinez-Seras et al. relied on the distance from an input latent vector to the set of class representatives, each defined by the centroid of the precomputed class latent distribution. However, reducing the geometry of a latent space to its centroid can be too restrictive when working with non-trivial connected latent manifolds or separated manifolds. In such cases, a single point may not serve as an adequate measure of distance from an arbitrary point in the space with respect to the class distribution. For example, when the manifolds are not connected, these centroids often lie outside the respective parts of the space.

To mitigate this effect, this work replaces the distance to the centroid with a point-set distance between the latent vector \( \bm{\ell} \in \mathbb{R}^N\) of a sample \( x \) and a set of latent vectors \(\mathcal{L}_c = \{\bm{\ell}^1, \bm{\ell}^2, \dots\}\) extracted from the complete latent manifold of class \( c \), where \( N \) is the dimension of the layer. This distance function is formulated using a prior distance function \( d(\bm{z}_1, \bm{z}_2) \), which measures the distance between two arbitrary vectors $z_1,z_2 \in \mathbb{R}^n$. The distance between a point and a set is then defined as the minimum distance between the point and any point in the set. From a theoretical perspective, this function should operate over the complete latent manifold \(\overline{\mathcal{L}_c} \subset \mathbb{R}^N \), however, due to the discrete nature of current AI, these clusters can only consist of a finite set of points. Therefore, we define the \textit{empirical distance} \(\widehat{d}(\cdot)\) by approximating the real manifold following the expression:
\begin{equation}
d(\bm{\ell}, \overline{\mathcal{L}_c}) = \inf_{\bm{\ell}_t \in \overline{\mathcal{L}}_c} d(\bm{\ell}, \bm{\ell}_t) \approx \min_{\bm{\ell}_t \in \mathcal{L}_c} d(\bm{\ell}, \bm{\ell}_t) = \widehat{d}(\bm{\ell}, \mathcal{L}_c) .
\end{equation}

Given that FFA provides different representations of the same input sample \( x \) depending on the embedded label \( p \) used during the supervised inference process, the input data used to create the latent space representatives must include both positive and negative instances of input data. Therefore, creating the set \(\mathcal{L}\) of the different clusters must involve a uniform amount of inputs from each class, using all possible labels. Once all these vectors have been captured, the set \(\mathcal{L}\) will provide a sufficient representation of the original latent space. Nevertheless, given that networks trained with FFA do not currently achieve state-of-the-art accuracy, a filtering step is necessary to remove potentially misclassified latent vectors. This step involves discarding a small number of the least active elements from sets where the embedded label aligns with the real class, as they may have been misidentified as negative samples. Conversely, it also involves removing a reduced set of the latent vectors with the highest activity from the negative sample set to avoid potential misclassification of these highly active elements as positive samples. To account for the presence of negative vectors, we extend the previous definition of \(\mathcal{L}\) for \(\{\mathcal{L}_{c, p} \mid c \in \mathcal{C} \text{ and } p \in \mathcal{C}\}\), where \(\mathcal{L}_{c,p}\) represents the set of latent vectors from class \( c \) forwarded using the embedded label \( p \), and \(\mathcal{C}\) denotes the set of class labels.

Once the representative set of latent vectors has been identified, an initial version of the scoring function can be formulated as the distance between the sample's latent vector and each class-representative set of clusters. Unlike conventional neural networks, where the latent representation of a sample is unique, FFA generates different representations depending on the embedded label used during the forward process. Therefore, to measure the distance between a sample and the set of class-representative clusters, it's crucial to account for the different representations that the sample obtains with different labels. Thus, the distance from the latent vector to each class is defined as the sum of the distances from the vector to each of the class clusters. This effect is illustrated in Figure \ref{fig:attrb}, where the green lines of the ID sample define the distance from each latent vector of the different embeddable labels with respect to the latent representative sets $\mathcal{L}_{c,p}$.

Consequently, the score of the unknown sample \(x\)is defined as the minimum distance between its latent representation, using the different possible input labels, and each set of class-representative clusters. Once this distance is computed, ID samples are expected to achieve a low overall distance, as their representation should remain close to their respective class set of clusters. In contrast, OoD samples are expected to attain a higher minimal distance, as their latent vectors should not be close to any set of ID clusters. Thus, the initial score function can be defined as
\begin{equation}
\label{eq:initial_ood}
    s(x) = \min_{p \in \mathcal{C}} \sum_{c \in \mathcal{C}} \widehat{d}\left( \mathcal{N}_k(x, c), \mathcal{L}_{p,c}\right)^\beta,
\end{equation}

where $\mathcal{N}_k(x, p)$ represents the output latent vector obtained from the forward pass of an input $x$ with the embedded label $p$ at depth $k$, and $\beta$ denotes a scaling hyperparameter.  

When working with certain spaces, this scoring function can encounter significant limitations, as an insufficient number of samples can lead to inaccuracies in the discrete approximation of the distance. For instance, in scenarios where the latent representation covers a large volume with low density, the large distance between points may cause the distance function to fail in finding any close neighbors, resulting in a high distance value. This issue is especially pronounced in high-dimensional spaces due to the curse of dimensionality, as evenly distributed representations require an exponential number of samples.

In such scenarios, an extreme outcome can occur where the scoring of samples is inverted, leading to OoD samples achieving lower scores than ID samples. This situation arises when the minimum distance from the zero vector to any representative cluster is smaller than the distance between points within the same representative cluster. Consequently, OoD samples, typically characterized by nearly zero latent activity, can end up with lower scores than ID samples with high latent activity, simply because they are unable to find a nearby neighbor within their own cluster to minimize the scoring function. Formally, this issue arises when the following inequality holds:
\begin{equation}
\label{eq:initial_ood_cond}
\min_{c \in C} \widehat{d}(\textbf{0}, L_{c,c} ) \le \mathbb{E}\left[\min_{c \in C} \widehat{d}(\mathcal{N}_p(\bm{X}, c), \mathcal{L}_{c,c} )\right],
\end{equation}
where $\bm{X}$ represents a randomly selected sample in the ID dataset. A more detailed explanation of this result is presented in \ref{ap:proof_equation}.

To mitigate this issue, an additional branch is incorporated into the original function to handle samples that meet this condition separately. Within this branch, instances closer to the zero vector than to any other class cluster are assigned a large value \( \gamma \). This high value indicates that these instances are more likely to belong to the negative distribution, which consists of synthetic samples outside the training distribution. However, we acknowledge that a small proportion of ID samples may also fall into this category. For these instances, a greater distance from the zero vector suggests a higher likelihood of misclassification. Therefore, we introduce a negative \( s(x) \) term into the expression, ensuring that elements that fall into this branch have their order reversed.

Therefore, after the proposed changes, the final scoring function is defined by:
\begin{equation}
    S(x) = \begin{cases}
        \gamma - s(x),&\text{if } s(x) > z_s \cdot s_0(x) \\
        s(x),&\text{otherwise,}
    \end{cases}
    \label{eq:ood_final_eq}
\end{equation}
where $\gamma$ denotes a constant value designed to assign an elevated score to samples falling into the first branch, the term $s_0(x)$ represents is equivalent to the function $s(\cdot)$ but replacing the spaces $\mathcal{L}_{i,j}$ for the space comprised by the vector $\bm{0}$, and $z_s$ serves as a scaling factor for the zero distance $s_0$. This final heuristic used with FF-OoD is detailed in Algorithm \ref{algo:ff_scp}.

\begin{algorithm}
\caption{FF-OoD}
\label{algo:ff_scp}
\begin{algorithmic}[1]
\STATE \textbf{INITIALIZE\_LATENTS}($X$, $Y$):
{\STATE \hspace{0.5cm}} \textbf{for} each $(x, y)$ in $(X, Y)$:
{\STATE \hspace{1cm}} \textbf{for} each $p$ in $\mathcal{C}$:
{\STATE \hspace{1.5cm}} Obtain the latent vector $\ell$ from $\mathcal{N}_1(x, p)$
{\STATE \hspace{1.5cm}} Append $\ell$ into the set $\mathcal{L}_{y, p}$

{\STATE \hspace{0.5cm}} \textbf{for} $(p,c)$ in $\mathcal{C} \times \mathcal{C}$
{\STATE \hspace{1cm}} \textbf{if} $p=c$
{\STATE \hspace{1.5cm}} Remove samples of $\mathcal{L}_{p, c}$ with low norm
{\STATE \hspace{1cm}} \textbf{else}
{\STATE \hspace{1.5cm}} Remove samples of $\mathcal{L}_{p, c}$ with high norm

\STATE \textbf{GET\_SCORE}($x$):
{\STATE \hspace{0.5cm}} Define the \textit{distance vector} $\bm{D}$ as $|\mathcal{C}|$ zeros
{\STATE \hspace{0.5cm}} Define the \textit{zero distance scalar} $D^0$ to zero
{\STATE \hspace{0.5cm}} \textbf{for} each $c$ in $\mathcal{C}$:
{\STATE \hspace{1cm}} Get the latent vector $\bm{\ell}$ from $\mathcal{N}_1(x, c)$
{\STATE \hspace{1cm}} Add the value $d(\bm{\ell}, \bm{0})$ to $D^0$
{\STATE \hspace{1cm}} \textbf{for} $p$ in $\mathcal{C}$:
{\STATE \hspace{1.5cm}} Add $\widehat{d}(\bm{\ell}, \mathcal{L}_{c,p})$ to $D_{p}$
{\STATE \hspace{0.5cm}} Let the score $s$ be the minimal value in the vector $\bm{D}$
{\STATE \hspace{0.5cm}} \textbf{return} $s$ \textbf{if} $s \geq D^0$ \textbf{else}  $\gamma - s$

\end{algorithmic}
\end{algorithm}

\subsection{Interpretation of Latent Vector}
\label{sec:approach_attribution_technique}

This section introduces a novel interpretability algorithm aimed at producing human-understandable explanations for the decisions made by FF-OoD. Given an input sample \( x \), this algorithm generates an attribution map highlighting the features that do not align with those expected from the data distribution of a selected class \( c \). To do so, it relies on the same latent space discrete approximation used by FF-OoD, which accurately approximates the emerging topology from networks trained using FFA (see \texttt{INITIALIZE\_LATENTS} in Algorithm \ref{algo:ff_scp}). By exploring this latent space, our algorithm identifies the closest neighbor \( x_f \) of the sample \( x \), ensuring that its latent representations closely resemble those expected from the latent data of class \( c \) while retaining key features that characterize the original input \( x \). A visual representation of this process is illustrated in Figure \ref{fig:attrb}.

The search of the latent space for the desired ID sample \( x_f \) is achieved by employing a decoder network that maps the latent space back to its pre-image space in \( D_{\oplus} \). Given the highly representational latent space of networks trained under FFA \citep{ororbia2023predictive}, we can assert that latent vectors surrounding the latent class manifold will result in plausible reconstructions, not distant from those present in the training dataset. We denote the application of this decoder network as \(\mathcal{N}^{-1} : \mathbb{L} \rightarrow D_{\oplus}\), which is trained using the input-latent pairs from the positive dataset. It's important to note that while negative samples belong to the latent space of the network, their features do not align with those expected from ID data. Therefore, to avoid deviating from ID classes, we operate solely within \( D_{\oplus} \).

This decoder network allows the algorithm to explore the latent space to find a distinct \( x_f \), derived from the input \( x \), that better aligns with the input distribution of a user-selected class \( c \in \mathcal{C} \). This exploration follows an optimization process in which a parameter latent vector \( \ell_w \) is optimized using gradient descent to minimize two key distances: a \textit{Latent Distance}, which measures the displacement between the parameter latent \( \ell_w \) and the manifold representing the latent cluster \( \mathcal{L}_{c,c} \); and a \textit{Decoder Distance}, which measures the distance between the pre-image \( x_f \) and the original sample \( x \). This loss function is encapsulated in the following expression:
\begin{equation}
L_{\textup{Attr}}(\ell_w; x, \alpha, \mathcal{L}_{c,c}) = 
\underbrace{
\|\mathcal{N}^{-1}(\ell_w) - x\|_2 \vphantom{d(\ell_w, \mathcal{L}_{c,c})}
}_{\text{Decoder Distance}}
+ 
   \underbrace{
\alpha \cdot \widehat{d}(\ell_w, \mathcal{L}_{c,c})
}_{\text{Latent Distance}},
\end{equation}

where \(\alpha\) denotes a hyperparameter used to balance the strength of each distance. Low values of \(\alpha\) result in latent values closer to the original image, which are desirable in cases where the sample is close to the desired distribution. Conversely, higher values of \(\alpha\) result in pre-images of the latent vector being closer to the input distribution of the target class, which is crucial when the distortion of the input data is large.

Once the optimization process is finished and the latent vector \( \ell_w \) has reached an equilibrium point between the desired targets, the attribution map is obtained by taking the difference between the original sample \( x \) and the optimal instance \( x_f \). This final heuristic has been detailed in Algorithm \ref{algo:attribution}.

\begin{algorithm}
\caption{Attribution Map Generation}
\label{algo:attribution}
\begin{algorithmic}[1]
\STATE \textbf{GET\_ATTRIBUTION}($x$, $c$, $\alpha$)
{\STATE \hspace{0.5cm}} Initialize the parameter latent vector $\bm{\ell}_w$ to $\mathcal{N}(x, c)$.
{\STATE \hspace{0.5cm}} \textbf{while} $\ell_w$ \textup{has not converged:}
{\STATE \hspace{1cm}} Compute the distance $\|\mathcal{N}^{-1}(\ell_w) - x\|_2$
{\STATE \hspace{1cm}} Compute the distance $\alpha \cdot \widehat{d}(\ell_w, \mathcal{L}_{c,c})$
{\STATE \hspace{1cm}} Obtain $L_{\text{Attr}}$ as the sum of both distances
{\STATE \hspace{1cm}} Update $\bm{\ell}_w$ with  a gradient descent step on $\nabla L_{\text{Attr}}$
{\STATE \hspace{0.5cm}} \textbf{return} $x - \mathcal{N}^{-1}(\bm{\ell}_w)$

\end{algorithmic}
\end{algorithm}

\section{Experimental Setup}\label{sec:exp_setup}
To systematically evaluate the methods presented in this work, we outline 3 Research Questions (RQs) that encapsulate the conditions under which our initial hypothesis would be satisfied, thereby establishing the practical utility of FFA. The chosen research questions are:\begin{itemize}
\item \textbf{RQ1}: Does the surrogate gradient result in competitive accuracy when training spiking FFA?
\item \textbf{RQ2}: Does the enhanced latent representation of spiking FFA, utilized by our proposed FF-OoD, improve upon the baseline results of SCP and ODIN?
\item \textbf{RQ3}: Does the latent space of models trained with spiking FFA exhibit sufficiently representative properties to develop interpretability mechanisms?
\end{itemize}

\subsection{Architecture and Training Process} \label{sec:exp_setup_training}

To conduct an unbiased experimental analysis across different network architectures and minimize potential biases between analog and spiking networks, all models were designed to exhibit equivalent behaviors. The first set of experiments was conducted on models consisting of two densely connected feedforward layers, each with 1400 neurons. Given the lack of competitive convolutional network implementations using FFA, the second set of experiments utilized a ResNet18 backbone to extract initial features \citep{he2016deep}, which were then passed to a two-layer fully-spiking network with $512$ neurons per layer. Analog models employed ReLU activation functions, while spike-based networks used LIF neurons as their biological counterpart. To implement these spiking dynamics, we used the \texttt{LIF} class from the \textit{snnTorch} \citep{eshraghian2021training} library, with the following hyperparameters: a \textit{fast sigmoid} as the surrogate function, $20$ timesteps per forward pass; a threshold of $0.4$ for the first layer and $0.3$ for the subsequent layer; and $\beta_n$ values of $0.4$ for the first layer and $0.2$ for the subsequent layer. To prevent spiking neurons from entering over-excitepd states, an input downscaling technique was applied, where each neuron’s input was scaled by a factor of $0.85^{N_{\text{Spikes}}}$ at each timestep, with $N_\text{Spikes}$ representing the total number of output spikes from that neuron at that moment.

The training process for all models followed the supervised methodology proposed by \citet{hinton2022forward}, wherein positive and negative samples are generated by embedding an encoded version of the label into the data. In addition to Hinton's proposed negative generation process, we also employed a greedy negative generation method, where the training negative label is chosen to the one showcasing the maximal goodness score. Each method was independently used to train each network, and the one resulting in higher accuracy was selected for further experimentation to ensure optimal predictive capabilities for subsequent tasks, including OoD detection and our attribution method. Additionally, we adopted the label initialization and embedding techniques proposed by \citet{lee2023symba}, where embedding vectors of class labels are set as random binary vectors appended at the end of the data. Each binary vector, of dimension 100, was initialized using a Bernoulli distribution with a probability of 0.1 for each element of the vector.

Networks without a backbone were trained for 10 epochs with a batch size of 512, using the hyperparameterized Binary Cross Entropy loss function outlined in Equation \ref{eq:probability_new}. Spiking networks were trained with the goodness functions \( G_{0}(\cdot) \) and \( G_{\infty}(\cdot) \). For networks utilizing the unbounded goodness function, the parameters were set to \( \theta^{+} = 6 \) and \( \theta^{-} = 2 \). Networks employing the bounded goodness function used thresholds of \( \theta^{+} = 0.3 \) and \( \theta^{-} = 0.1 \), along with \( \alpha = \beta = 5 \) to optimize the probability function’s scaling. Training was performed using the ADAM optimizer with a learning rate of 0.001 for ANNs and 0.002 for SNNs. In contrast, networks using a ResNet backbone were trained for 100 epochs. For unbounded models, the parameters were set to $\theta=12$ with $\alpha$ and $\beta$ both equal to 5. For bounded models, $\theta$ values were set to 0.1 for CIFAR-10 and 0.5 for SVHN, with $\alpha$ and $\beta$ set to 10 for CIFAR-10 and 4 for SVHN. These values were determined through early experimentation and manual tuning. All the code has been uploaded into \url{https://github.com/AnonymousSquirrel316/FFA_OOD}.

\subsection{Out-of-Distribution Evaluation Setup} \label{sec:exp_setup_ood}
As this paper builds upon the groundwork of the SCP algorithm introduced by \citet{seras2022novel}, and given the absence of subsequent spiking-specific OoD detection methods in the literature, the benchmarks used for this work are extracted from the aforementioned study. Specifically, the SCP and ODIN algorithms stand as the top-performing approaches, serving as the designated benchmarks for comparative analysis in this work. Furthermore, to demonstrate the competitiveness of our algorithm in the spiking domain compared to its analog counterpart, both types of networks have been tested.

The models used in these experiments were the best-performing ones from the training phase of RQ1. During early experimentation, we explored Euclidean, Manhattan, and Cosine distances for the distance function \( d(\cdot, \cdot) \) and selected Manhattan distance for non-backbone networks and Euclidean distance for networks using a feature extractor, as they yielded superior performance. To construct the latent space, we used 1024 samples for networks without a backbone and 5000 samples for networks with a backbone, following the procedure outlined in the \texttt{initialize\_latents} function of Algorithm \ref{algo:ff_scp}. For the filtering step, we removed 20\% of the highest and lowest latent vectors from their respective latent sets. The hyperparameters \(\gamma\), \(\beta\), and \(z_0\) were selected through an early hyperparameter selection process for each training and goodness. Their search space was defined as follows: \(z_0\) was chosen from the set \(\{0.85, 1, 1.15, 1.4\}\); \(\gamma\) was either \(10^4\) or \(10^6\); and \(\beta\) was set to either 1 or 2.

To objectively evaluate the model's OoD detection performance across the experiments, three key evaluation metrics have been measured: AUROC, AUPR, and FPR95. These metrics are widely employed in the OoD detection literature, providing valuable insights into the model's ability to distinguish between in-distribution and out-of-distribution samples. They are defined as:

\begin{enumerate}
    \item \textbf{AUROC (Area Under the Receiver Operating Characteristic curve)}: Quantifies the model's ability to discriminate between in-distribution and out-of-distribution samples. A higher AUROC indicates better separation, with 1.0 representing perfect discrimination and 0.5 indicating random guessing.
    \item \textbf{AUPR (Area Under the Precision-Recall curve)}: Evaluates the trade-off between precision and recall for detecting OoD samples. It is particularly useful when dealing with imbalanced datasets, as it emphasizes the model's ability to correctly identify OoD instances.
    \item \textbf{FPR95 (False Positive Rate at 95\% True Positive Rate)}: Measures the false positive rate when the true positive rate is fixed at 95\%. Lower FPR95 scores indicate higher OoD detection performance.
\end{enumerate}

\subsection{Evaluation of the Quality of the Attribution Method} \label{sec:exp_setup_attrib}

As mentioned in Section \ref{sec:approach_attribution_technique}, our proposed interpretability algorithm provides a visual representation of the areas that drive images towards the OoD direction. Given the subjective nature of evaluating the quality of an attribution map, we rely on a qualitative analysis of several samples from the training data. We conducted this analysis using instances from the MNIST and KMNIST datasets, generating altered images through several obstruction methods to measure our method's performance when presented with closely resembling images. The artifacts introduced in these images serve as primary factors pushing them towards the OoD direction, and thus, should be the most prominent regions highlighted on the attribution maps. The samples chosen for evaluation were randomly selected from both datasets, and after comprehensive inspection, those that most clearly embody the characteristics captured by the attribution map were retained for further analysis. This analysis focuses on identifying the attribution capabilities of the approach and uncovering potential limitations.

The decoder network chosen to reconstruct the latent spaces is a single-layer analog network utilizing Sigmoid activations. To prepare the spike train vectors for the analog decoder, these latent vectors are averaged over the time dimension. During training, the decoder minimizes Binary Cross Entropy loss between the input image and the reconstructed image generated from the latent vector obtained from the first layer. This loss function was preferred due to its ability to produce more accurate results compared to regression-based losses. Similarly, we chose to extract latent vectors from the first layer because of its accurate reconstructions. The decoder was trained for 50 epochs using the ADAM optimizer with a batch size of 512 and an initial learning rate of 0.001. Additionally, an exponential decay on the learning rate was applied, with a gamma decay of 0.4 every 5 epochs. The value of \(\alpha\) has been manually tuned in each experiment, and the one resulting in the best attribution map has been selected.

\begin{wrapfigure}{r}{0.4\textwidth} 
    \centering
    \includegraphics[width=0.37\columnwidth]{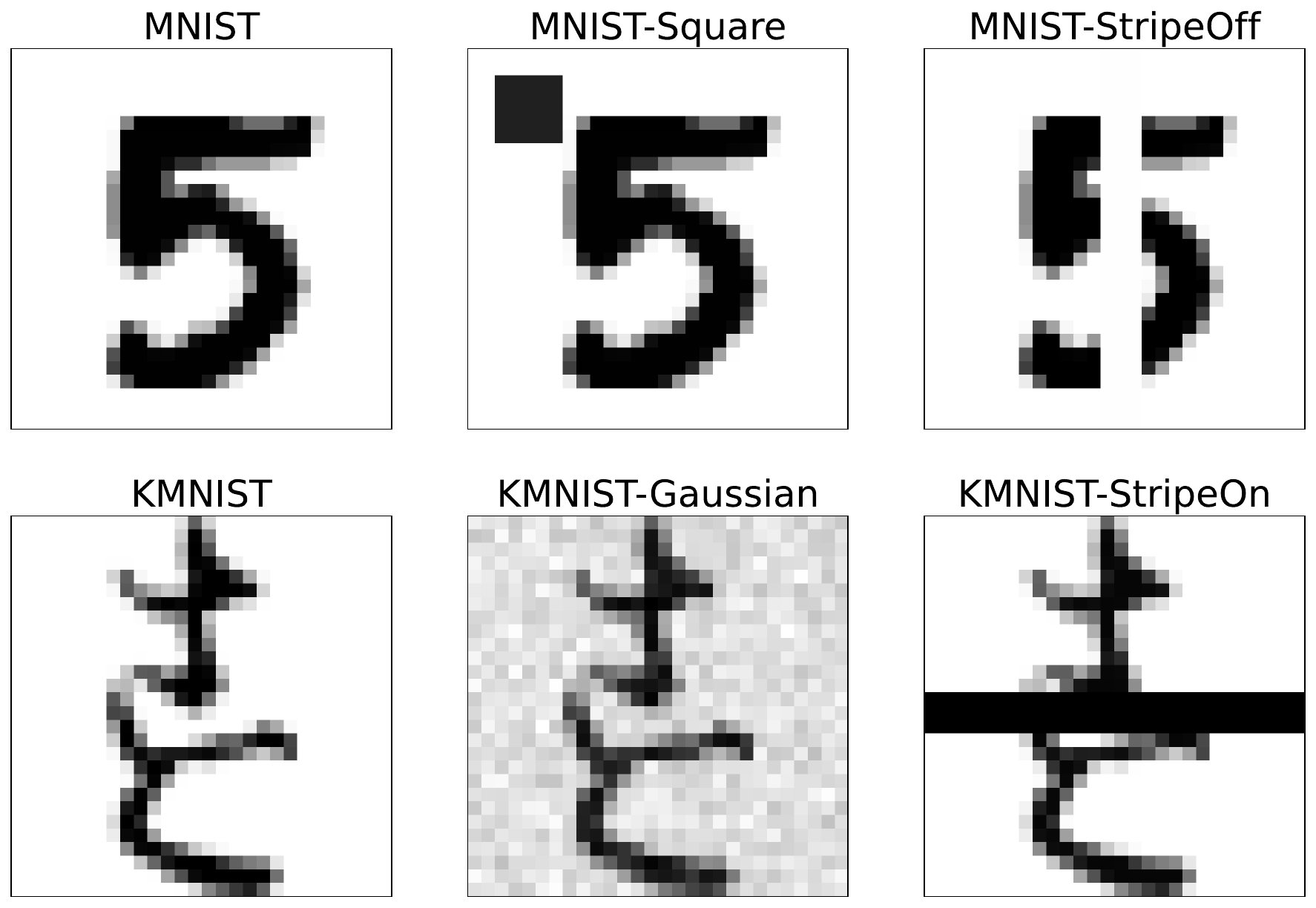}
    \caption{MNIST and KMNIST ID samples followed by OoD version obtained by applying one obstruction method.}
    \label{fig:attr_ood_examples}
\end{wrapfigure}

The obstruction techniques employed in these experiments include: \textit{Square}, which introduces a 5x5 square at a random corner of the image, positioned 2 pixels away from the borders of one of the corners; \textit{Gaussian}, which injects random Gaussian noise into the image, drawn from the normal distribution $\mathcal{N}(0, 0.2)$; \textit{StripeOff}, which eliminates a random vertical or horizontal stripe of 3 pixels from the image, with the removal occurring within the center (pixels from 12th to 18th row or column); and \textit{StripeOn}, which adds a random vertical or horizontal stripe of 3 pixels to the image within the center (pixels from 12th to 18th row or column). All proposed obstruction methods are illustrated in Figure \ref{fig:attr_ood_examples}.

\subsection{In-Distribution and Out-of-Distribution Datasets} \label{sec:exp_setup_datasets}
To evaluate the OoD Detection capacity of spiking networks, we trained the networks in 5 ID datasets for the feedforward (MNIST \citep{lecun_mnist_2010}, Fashion MNIST (FMNIST) \citep{xiao2017fashion}, Kuzushiji-MNIST (KMNIST) \citep{clanuwat2018deep}, and EMNIST (only the Letters subset) \citep{cohen2017emnist}) and 2 for the convolutional alternative (CIFAR10 \citep{krizhevsky2009learning} and SVHN \citep{netzer2011reading}). In these experiments, the OoD dataset is constructed by combining the datasets from each category excluding the one used for the training. To expand the scope of our analysis, we introduce two additional OoD datasets for the densely connected model (Not-MNIST (small) \citep{bulatov2011notmnist} and Omniglot \citep{lake2015human}) and six for the convolutional (Food101 \citep{bossard14}, Flowers102 \citep{nilsback2008automated}, Caltech101 \citep{fei2006one}, GTSRB \citep{stallkamp2011german}, EuroSAT \citep{helber2019eurosat} and DTD \citep{cimpoi2014describing}). For the densely connected model, a uniform normalization pipeline was applied across all datasets and all images were transformed into 20 timesteps of uniform activity. Similarly, the outputs of the ResNet18 backbone followed the same transformation before going into networks trained using FFA.

\section{Results and Discussion}\label{sec:res}

This section presents the results from various experiments, discussing the findings and highlighting observed limitations. Initially, Section \ref{sec:train_results} showcases the accuracy results of our proposed spiking adaptations of FFA. Section \ref{sec:ood_results} presents the outcomes of the OoD detection task using the FF-OoD algorithm, comparing it against the SCP and ODIN algorithms. Subsequently, Section \ref{sec:attribution_results} discusses the results obtained from the attribution method proposed in this paper. Finally, Section \ref{sec:limitations} discusses the limitations observed during experimentation.

\subsection{RQ1: Training Results of spiking FFA}
\label{sec:train_results}

The accuracies obtained from the training experiments using the different goodness functions are presented in Table \ref{tab:Accuracy_ANN_SNN}. These results demonstrate the comparable performance between the analog and the spiking implementations of FFA, where SNNs only exhibit a slight decrease in accuracy, likely due to the increased noise within latent vectors caused by the stochastic input preparation process and the information loss due to the discretization of the latent states. Between the two proposed goodness functions, the models employing the unbounded version achieved higher accuracy than their bounded counterparts, probably as a consequence of the probability function not needing additional scaling parameters to reach the extreme values. 

Nevertheless, in non-convolutional networks, while the average accuracy shows a noticeable drop, this loss in accuracy is primarily related to the significant difference in accuracy on the KMNIST dataset, where the bounded version reduces accuracy by \(14.61\) points. This abnormal relationship in the drop in accuracy across different datasets suggests that the bounded and unbounded models process the data differently, leading to distinct classification heuristics and representation properties. These results align with the observation that networks using bounded goodness functions do not experimentally create sparse latent representations, which can result in reduced feature specificity in the neurons (see \ref{ap:latent_geometry}).

When examining networks trained using the output of a feature extractor, the results show that only unbounded goodness remains competitive. The highly variable activity distribution in the feature extractor's latent space leads to goodness scores that fail to adequately cover the domain of the probability function. This results in suboptimal gradient updates, which eventually converge to less accurate parameters. As a result, these experiments appear to have an accuracy decrease of approximately a 10\%.

\begin{table}[h!]
  \centering
  \small
  \caption{Accuracy of networks trained with the Forward-Forward Algorithm (FFA) using an analog and two spiking implementations. The spiking versions employ the $G_0(\cdot)$ and $G_{\infty}(\cdot)$ goodness functions respectively.}
  \vspace{10pt}
  \begin{tabular}{ll>{\centering\arraybackslash}m{2.4cm}>{\centering\arraybackslash}m{2.4cm}>{\centering\arraybackslash}m{2.4cm}>{\centering\arraybackslash}m{2.4cm}}
    \toprule[2pt]
        \midrule
    Network & Dataset & FFA \hspace{1cm}{\scriptsize Analog} &  FFA \hspace{1cm}{\scriptsize Spiking $G_0(\cdot)$} &  FFA \hspace{1cm}{\scriptsize Spiking $G_{\infty}(\cdot)$} \\
    \midrule[1pt]
    \multirow{4}{*}{Dense} & MNIST &  97.77\% & 93.27\% & 93.01\% \\ 
     &FMNIST &  85.75\% & 83.69\% & 85.68\% \\ 
     &KMNIST & 91.86\% &  71.30\% &  85.91\% \\ 
     &EMNIST &   87.25\% & 76.54\% & 80.07\% \\
    \midrule
    \multirow{2}{*}{Image} & CIFAR10 & 69.86\% & 61.7\%& 69.60\%\\
     & SVHN  &  90.25\%& 81.78\%& 90.57\% \\
    \bottomrule
  \end{tabular}  
  \label{tab:Accuracy_ANN_SNN}
\end{table}

\subsection{RQ2: Out-of-Distribution Detection Results}
\label{sec:ood_results}

The metrics of the OoD detection experiments using our proposed method are presented in Table \ref{tab:snn_ood_results_comp_final} for dense models and in Table \ref{tab:snn_ood_results_comp_final_visual} for image data. After thoroughly analyzing the results, we conclude that they have yielded a positive outcome to our initial research question. However, we remark the current limitations of FFA when dealing with image data, which we analyze separately from the fully FFA-based implementation. A detailed discussion of these results follows in the present subsection.

\begin{table*}
    \definecolor{mygray}{RGB}{80,80,80}
    \newcommand{\ceco}{\cellcolor{blue!6}}
    \newcommand{\teco}[1]{\textbf{\textcolor{mygray}{#1}}}
    \centering
    \caption{Results of the FF-OoD algorithm of our three proposed networks: the Analog FF-Network; the Unbounded FF-Network, denoted as $G_\infty(\cdot)$; and the Bounded FF-Network, denoted as $G_0(\cdot)$. Data from the SCP and ODIN has been extracted from the paper \cite{seras2022novel}. The ANN results are only shown as a baseline to compare ANN vs SNN.}
    \label{tab:snn_ood_results_comp_final}
    
    \resizebox{\textwidth}{!}{
    \begin{tabular}{clcccccp{0.2cm}cccccp{0.2cm}ccccc}
        \toprule[2pt]
        \midrule
        & & \multicolumn{5}{c}{AUROC $\uparrow$}  & & \multicolumn{5}{c}{ AUPR $\uparrow$}    & & \multicolumn{5}{c}{FPR95 $\downarrow$}      \\ \cmidrule{3-7} \cmidrule{9-13} \cmidrule{15-19}   
        \makecell[c]{ID\\Dataset} & \makecell[c]{OOD\\Dataset} &    \thead{FF-OoD \\ {\small Analog}} &  \thead{FF-OoD \\ {\small $G_{\infty}(\cdot)$}} &  \thead{FF-OoD \\ {\small $G_{0}(\cdot)$}}  & SCP & ODIN   & &  \thead{FF-OoD \\ {\small Analog}} &  \thead{FF-OoD \\ {\small $G_{\infty}(\cdot)$}} &  \thead{FF-OoD \\ {\small $G_{0}(\cdot)$}} & SCP & ODIN  & & \thead{FF-OoD \\ {\small Analog}} & \thead{FF-OoD \\ {\small $G_{\infty}(\cdot)$}} & \thead{FF-OoD \\ {\small $G_{0}(\cdot)$}} & SCP & ODIN    \\       
        \midrule
         \multirow{5}{*}{MNIST}
         & Omniglot   & \teco{91.14} & 90.18 & 89.07& 71.56  & \textbf{94.72} & & \teco{94.41} & 92.12 & 91.87& 71.90 & \textbf{94.28} & & \teco{37.18} & 29.97 & 34.38& 88.93 & \textbf{29.08} \\
         & notMNIST   & \teco{\textbf{99.99}} & \textbf{99.97} & 99.97& 99.14  & 64.45 & & \teco{99.99} & 99.94 & \textbf{99.99}& 99.01 & 57.16 & & \teco{\textbf{0.00}} & 0.01 & \textbf{0.00} & 03.60 & 86.65 \\
         & FMNIST     & \teco{97.97} & 97.87 & \textbf{99.37}& 91.21  & 73.97 & & \teco{97.87} & 97.71 & \textbf{99.40}& 87.15 & 65.29 & & \teco{10.54} & 11.72 & \textbf{3.34}& 24.95 & 77.11 \\
         & KMNIST     & \teco{98.12} & 98.40 & \textbf{99.39}& 94.89  & 73.97 & & \teco{98.05} & 98.33 & \textbf{99.41}& 94.92 & 76.27 & & \teco{8.72} & 8.88 & \textbf{3.46}& 24.99 & 72.87   \\
         & Letters    & \teco{89.63} & 91.60 & \textbf{96.85}& 88.11  & 81.43 & & \teco{94.62} & 95.72 & \textbf{98.45}& 79.49 & 64.19 & & \teco{54.62} & 40.12 & \textbf{16.00}& 51.66 & 73.89 \\
        \midrule
        \multirow{5}{*}{FMNIST}
        & MNIST       & \teco{99.10} & \textbf{99.00} & 98.76& 92.60  & 95.98 & & \teco{98.50} & 98.12 & \textbf{98.23}& 93.18 & 95.74 & & \teco{3.14} & \textbf{3.44} & 4.64& 35.48 & 17.37   \\
         & Omniglot   & \teco{98.86} & \textbf{98.67} & 96.38& 84.75  & 97.86 & & \teco{98.85} & \textbf{98.61} & 97.40& 85.37 & 97.83 & & \teco{3.84} & \textbf{4.33} & 15.99& 63.55 & 09.97 \\
         & notMNIST   & \teco{99.85} & 99.89 & \textbf{99.91}& 98.87  & 61.70 & & \teco{99.93} & 99.75 & \textbf{99.95}& 99.14 & 61.75 & & \teco{0.21} & \textbf{0.12} & 0.33& 02.47 & 92.33 \\
         & KMNIST     & \teco{97.44} & \textbf{97.82} & \textbf{98.05}& 84.56  & 88.36 & & \teco{97.20} & 97.16 & \textbf{97.82} & 82.41 & 88.84 & & \teco{12.84} & 9.13 & \textbf{8.97}& 53.10 & 57.39 \\
         & Letters    & \teco{98.05} & \textbf{98.31} & 97.41& 92.83  & 91.83 & & \teco{98.75} & \textbf{98.69} & 98.65& 88.10 & 85.15 & & \teco{8.72} & \textbf{6.60} & 13.64& 34.49 & 39.82 \\
        \midrule
        \multirow{5}{*}{KMNIST}
         & MNIST      & \teco{86.26} & 81.93 & 60.73& 61.85 & \textbf{85.79} & & \teco{76.83} & 70.23 & 53.89& 61.64 & \textbf{88.57} & & \teco{36.11} & \textbf{44.11} & 75.78& 89.34 & 51.64   \\
         & Omniglot   & \teco{86.26} & 82.59 & 54.84& 58.54 & \textbf{96.43} & & \teco{86.84} & 84.12 & 62.58& 59.68 & \textbf{97.25} & & \teco{39.04} & 43.24 & 69.36& 92.58 & \textbf{08.09} \\
         & notMNIST   & \teco{95.95} & 91.52 & \textbf{99.13}& 96.88 & 67.24 & & \teco{98.03} & 88.48 & \textbf{99.53}& 96.68 & 67.59 & & \teco{28.91} & 18.34 & \textbf{3.89}& 09.52 & 81.37 \\
         & FMNIST     & \teco{84.18} & \textbf{77.45} & 67.84& 75.00 & 76.73 & & \teco{88.03} & \textbf{83.46} & 60.86& 72.85 & 78.58 & & \teco{79.82} & 80.68 & 71.89& \textbf{66.14} & 70.67 \\
         & Letters    & \teco{73.86} & 79.37 & 82.75& 78.34 & \textbf{85.14} & & \teco{79.95} & 82.38 & \textbf{84.89}& 64.47 & 80.65 & & \teco{59.83} & 48.55 & \textbf{44.96}& 71.49 & 53.51 \\
        \midrule
        \multirow{5}{*}{Letters}
        & MNIST       & \teco{88.04} & 83.36 & \textbf{83.93}& 69.06 & 80.98 & & \teco{81.16} & 76.13 & 71.91& 82.30 & \textbf{88.73} & & \teco{52.45} & 71.13 & \textbf{65.56}& 88.83 & 68.13   \\
         & Omniglot   & \teco{81.14} & 82.59 & 84.97& 77.60 & \textbf{97.11} & & \teco{78.24} & 78.51 & 77.37& 88.04 & \textbf{98.54} & & \teco{63.38} & 54.51 & 45.13& 87.11 & \textbf{15.54} \\
         & notMNIST   & \teco{99.98} & \textbf{99.97} & 99.94 & 98.63 & 58.08 & & \teco{99.98} & 99.81 & \textbf{99.94} & 98.90 & 69.65 & & \teco{\textbf{0.00}} & 0.02 & \textbf{0.00}& 07.48 & 93.94 \\
         & FMNIST     & \teco{91.62} & \textbf{95.94} & 90.78& 86.33 & 71.30 & & \teco{90.13} & \textbf{94.56} & 84.58& 92.08 & 83.66 & & \teco{60.02} & \textbf{27.63} & 37.70& 47.30 & 88.04\\
         & KMNIST     & \teco{99.27} & \textbf{99.61} & 97.47& 86.31 & 77.60 & & \teco{98.72} & \textbf{99.01} & 95.51& 92.15 & 85.47 & & \teco{3.47} & \textbf{1.44} & 14.16& 51.41 & 77.88 \\
        \midrule
         &  \textbf{Average} & \teco{92.84} & \textbf{92.30} & 89.88 & 84.35 & 81.03 & & \teco{92.80} & \textbf{91.64} & 88.61 & 84.47 & 81.25 & & \teco{28.14} & \textbf{25.20} & 26.46& 49.72 & 58.26 \\
         &  \textbf{Median} & \teco{96.70} & 96.88 & \textbf{97.13} & 86.32 & 81.20 & & \teco{97.54} & 96.44 & \textbf{97.61} & 87.59 & 84.40 & & \teco{20.88} & \textbf{15.03} & 15.07& 51.53 & 69.4 \\
        \bottomrule
        \end{tabular}}    
\end{table*}

\paragraph{sFF-OoD is competitive against SCP and ODIN} The main hypothesis guiding the development of the FF-OoD algorithm was that the improved representation capabilities of networks trained with FFA, together with refined geometric methods, would yield higher accuracy in OoD Detection tasks. The results, as presented in Table \ref{tab:snn_ood_results_comp_final} and Table \ref{tab:snn_ood_results_comp_final_visual}, verify this hypothesis, with both the bounded and unbounded models consistently outperforming prior methods in densely connected networks, and unbounded methods achieving higher AUROC's in models with a convolutional backbone.

Nevertheless, the FF-OoD algorithm appears to follow the same pattern of the best- and worst-performing pairs of ID and OoD datasets as the SCP algorithm, confirming the similarity between the heuristics of both methods. A plausible explanation for the reduced results on KMNIST as an ID dataset may arise from the suboptimal test performance exhibited by networks using FFA. Upon examining the weights they generate, they demonstrate less feature-selectivity, allowing for easier OoD samples to exhibit similar latent vectors to ID samples. Additionally, consistent with the findings of \citet{seras2022novel}, the Omniglot dataset remains the poorest performing as an OoD dataset, with ODIN outperforming both algorithms. However, both FF-OoD and SCP demonstrate the capability to outperform ODIN in the MNIST, FMNIST, and Letters datasets, reinforcing their potential in specific scenarios.

\paragraph{Bounded vs Unbounded Goodness in FF-OoD} Both goodness functions show remarkable similarity across most tasks in the benchmark, with the highest accuracy alternating depending on the specific task. However, the bounded goodness displays suboptimal accuracy in the KMNIST task, with numerous outcomes falling below 70 in terms of AUROC. This performance discrepancy may be attributed to the fact that the latent vectors of the bounded network lack sparsity, resulting in less specialization within the representation. This lack of neural specialization, together with the non-competitive accuracy on the dataset, may be explained by class clusters not achieving enough distance between them, thereby explaining the low accuracy when using the FF-OoD algorithm.
\paragraph{SNNs compete with ANNs when employing FF-OoD} To ensure that SNNs offer practical advantages over ANNs, it is essential to verify that their performance is not significantly overshadowed by their analog counterparts. Such a scenario would suggest that the architectural benefits of SNNs, such as inference speed and energy efficiency, might become secondary to accuracy advantages. The results presented in Table \ref{tab:snn_ood_results_comp_final} demonstrate how the FF-OoD employed with unbounded goodness on SNNs yields outcomes comparable to its analog counterpart. Both methods achieve AUROC accuracies with an average difference of less than one point. Analog networks using FF-OoD perform slightly better than SNNs in average AUROC and both average and median AUPR. Conversely, both SNN goodness functions achieve superior results in median AUROC and both FPR95 measures. Examining specific instances, the worst result in analog networks is approximately 5 points below the worst result in its unbounded spiking counterpart. The bounded SNN counterpart experiences an additional drop when presented with KMNIST as the ID dataset, primarily due to the previously mentioned lack of test accuracy. However, the overall difference in accuracy is minimal, with this bounded version achieving higher AUROC in multiple cases.

\begin{table*}
    \definecolor{mygray}{RGB}{80,80,80}
    \newcommand{\ceco}{\cellcolor{blue!6}}
    \newcommand{\teco}[1]{\textbf{\textcolor{mygray}{#1}}}
    \centering
    \caption{Results of the FF-OoD algorithm of our three proposed networks: the Analog FF-Network; the Unbounded FF-Network, denoted as $G_\infty(\cdot)$; and the Bounded FF-Network, denoted as $G_0(\cdot)$. Data from the SCP and ODIN has been extracted from the paper \cite{seras2022novel}. The ANN results are only shown as a baseline to compare ANN vs SNN.}
    \label{tab:snn_ood_results_comp_final_visual}
    
    \resizebox{\textwidth}{!}{
    \begin{tabular}{clcccccp{0.2cm}cccccp{0.2cm}ccccc}
        \toprule[2pt]
        \midrule
        & & \multicolumn{5}{c}{AUROC $\uparrow$}  & & \multicolumn{5}{c}{ AUPR $\uparrow$}    & & \multicolumn{5}{c}{FPR95 $\downarrow$}      \\ \cmidrule{3-7} \cmidrule{9-13} \cmidrule{15-19}   
        \makecell[c]{ID\\Dataset} & \makecell[c]{OOD\\Dataset} &    \thead{FF-OoD \\ {\small Analog}} &  \thead{FF-OoD \\ {\small $G_{\infty}(\cdot)$}} &  \thead{FF-OoD \\ {\small $G_{0}(\cdot)$}}  & SCP & ODIN   & &  \thead{FF-OoD \\ {\small Analog}} &  \thead{FF-OoD \\ {\small $G_{\infty}(\cdot)$}} &  \thead{FF-OoD \\ {\small $G_{0}(\cdot)$}} & SCP & ODIN  & & \thead{FF-OoD \\ {\small Analog}} & \thead{FF-OoD \\ {\small $G_{\infty}(\cdot)$}} & \thead{FF-OoD \\ {\small $G_{0}(\cdot)$}} & SCP & ODIN    \\       
        \midrule
         \multirow{7}{*}{CIFAR10}
         & Food101      & \teco{85.25} & \textbf{84.02} & 76.61 & 71.65 & 68.10 &   & \teco{91.39} & \textbf{91.27} & 86.17 & 70.30 & 70.12 &   & \teco{48.57} & \textbf{59.02} & 66.46 & 81.56 & 89.99\\
         & Flowers102   & \teco{86.88} & \textbf{85.64} & 79.44 & 75.77 & 68.71 &   & \teco{78.08} & \textbf{79.95} & 63.66 & 75.04 & 70.93 &   & \teco{48.57} & 60.61 & \textbf{52.63} & 75.59 & 90.54\\
         & Caltech101   & \teco{70.63} & \textbf{68.48} & 67.79 & 67.02 & 65.88 &   & \teco{65.94} & 63.76 & 60.46 & 64.47 & \textbf{67.90} &   & \teco{83.44} & \textbf{81.30} & 81.39 & 83.81 & 91.53\\
         & GTSRB        & \teco{89.76} & \textbf{88.82} & 81.27 & 78.76 & 71.81 &   & \teco{89.55} & \textbf{89.99} & 79.72 & 78.10 & 75.42 &   & \teco{28.08} & \textbf{50.69} & 52.21 & 69.49 & 91.38\\
         & SVHN         & \teco{89.25} & \textbf{86.28} & 77.74 & 62.70 & 70.10 &   & \teco{94.24} & \textbf{93.63} & 87.46 & 61.21 & 72.27 &   & \teco{31.69} & 62.34 & \textbf{62.23} & 88.51 & 89.41\\
         & EuroSAT      & \teco{90.27} & \textbf{87.98} & 75.10 & 65.88 & 70.00 &   & \teco{95.37} & \textbf{95.14} & 86.74 & 64.91 & 71.43 &   & \teco{33.98} & \textbf{62.41} & 63.49 & 85.69 & 88.96\\
         & DTD          & \teco{83.74} & \textbf{82.48} & 77.13 & 61.43 & 81.76 &   & \teco{45.80} & 53.86 & 33.43 & 63.91 & \textbf{85.12} &   & \teco{62.50} & 71.69 & \textbf{66.10} & 90.60 & 83.95\\
        \midrule
        \multirow{7}{*}{SVHN}
         & Food101      & \teco{81.48} & \textbf{85.59} & 70.89 & 61.73 & 63.97 &   & \teco{71.83} & \textbf{77.82} & 64.46 & 59.24 & 64.94 &   & \teco{45.73} & \textbf{36.38} & 76.56 & 84.89 & 92.71 \\
         & Flowers102   & \teco{83.66} & \textbf{87.24} & 72.86 & 80.47 & 77.92 &   & \teco{43.47} & 53.49 & 32.78 & 81.66 & \textbf{82.68} &   & \teco{36.64} & \textbf{33.81} & 68.88 & 60.79 & 93.20 \\
         & Caltech101   & \teco{79.19} & \textbf{83.54} & 69.22 & 80.94 & 77.24 &   & \teco{45.93} & 53.60 & 36.89 & \textbf{82.75} & 81.86 &   & \teco{57.67} & \textbf{47.18} & 77.83 & 60.35 & 91.97 \\
         & GTSRB        & \teco{80.72} & 81.04 & 56.30 & \textbf{81.82} & 77.02 &   & \teco{60.55} & 64.83 & 36.41 & \textbf{82.23} & 81.85 &   & \teco{56.61} & 67.03 & 87.21 & \textbf{56.24} & 90.09 \\
         & CIFAR10      & \teco{79.41} & \textbf{85.48} & 72.67 & 83.11 & 81.01 &   & \teco{48.17} & 57.88 & 43.61 & \textbf{83.80} & 81.10 &   & \teco{54.23} & \textbf{37.06} & 73.70 & 55.59 & 88.64 \\
         & EuroSAT      & \teco{82.49} & \textbf{84.18} & 73.49 & 79.06 & 77.73 &   & \teco{78.32} & \textbf{83.03} & 68.19 & 79.38 & 82.25 &   & \teco{53.10} & \textbf{60.00} & 63.13 & 60.84 & 92.32 \\
         & DTD          & \teco{77.47} & \textbf{82.28} & 70.60 & 81.16 & 77.69 &   & \teco{13.56} & 17.30 & 11.83 & \textbf{92.85} & 82.39 &   & \teco{55.90} & \textbf{43.73} & 73.83 & 56.75 & 81.61 \\
        \midrule
    &  \textbf{Average} & \teco{82.87} & \textbf{83.78} & 72.93 & 73.67 & 73.49 &   & \teco{65.87} & 69.68 & 56.55 & 74.27 & \textbf{76.44} &   & \teco{49.76} & \textbf{55.23} & 68.97 & 72.19 & 89.73 \\
    &  \textbf{Median}  & \teco{83.07} & \textbf{84.83} & 73.17 & 77.26 & 74.41 &   & \teco{68.88} & 71.32 & 62.06 & 76.57 & \textbf{78.26} &   & \teco{50.83} & \textbf{59.51} & 67.67 & 72.54 & 90.31 \\
        \bottomrule
        \end{tabular}}    
\end{table*}
\paragraph{Importance of the scoring $\mathbf{S(x)}$ over $\mathbf{s(x)}$} To empirically demonstrate the necessity of this modification, the results obtained on the same benchmarks using the FF-OoD algorithm with the scoring function \( s(x) \) are presented in Table \ref{tab:snn_ood_results_comp_s_x}, in \ref{ap:ood_different_sx}. In these results, the impact of this modification becomes evident when the ID dataset is KMNIST and the OoD dataset is FMNIST, where the AUROC drops significantly to 32.95 under models trained with the unbounded spiking FFA and to 17.92 for analog networks. However, when the condition specified in Equation \eqref{eq:hold_condition_aen} (namely, the low intra-class distance) does not hold, the algorithm remains capable of accurately detecting OoD samples. This is particularly observed in the bounded version of the goodness, where the latent space exhibits highly active latent vectors, consistently avoiding the reversal in the scoring values.

\paragraph{FFA requires good image backbones} In contrast to the fully dense model presented in Table \ref{tab:snn_ood_results_comp_final}, the model employing a convolutional backbone shows mixed results. As shown in Table \ref{tab:snn_ood_results_comp_final_visual}, its AUROC performance remains comparable to that of SCP and ODIN. However, the AUPR scores drop significantly when evaluated on the SVHN ID dataset. This discrepancy likely stems from the pretraining of ResNet18 on ImageNet, which shares greater similarity with CIFAR-10, another natural image dataset. In contrast, SVHN, which comprises images of numerical digits, differs significantly in distribution, leading to suboptimal AUPR performance compared to CIFAR-10. Nevertheless, the latent vectors contain enough feature information for FF-OoD to still high OoD Detection accuracy on instances where SVHN and OoD datasets are distinct enough (e.g. Food101 or EuroSAT).

\subsection{RQ3: Attribution Results}
\label{sec:attribution_results}

\begin{figure*}[t]
\centering
    \subfloat[\scriptsize MNIST-Square ($\alpha = 0.1$)]{
        \includegraphics[width=0.3\textwidth]{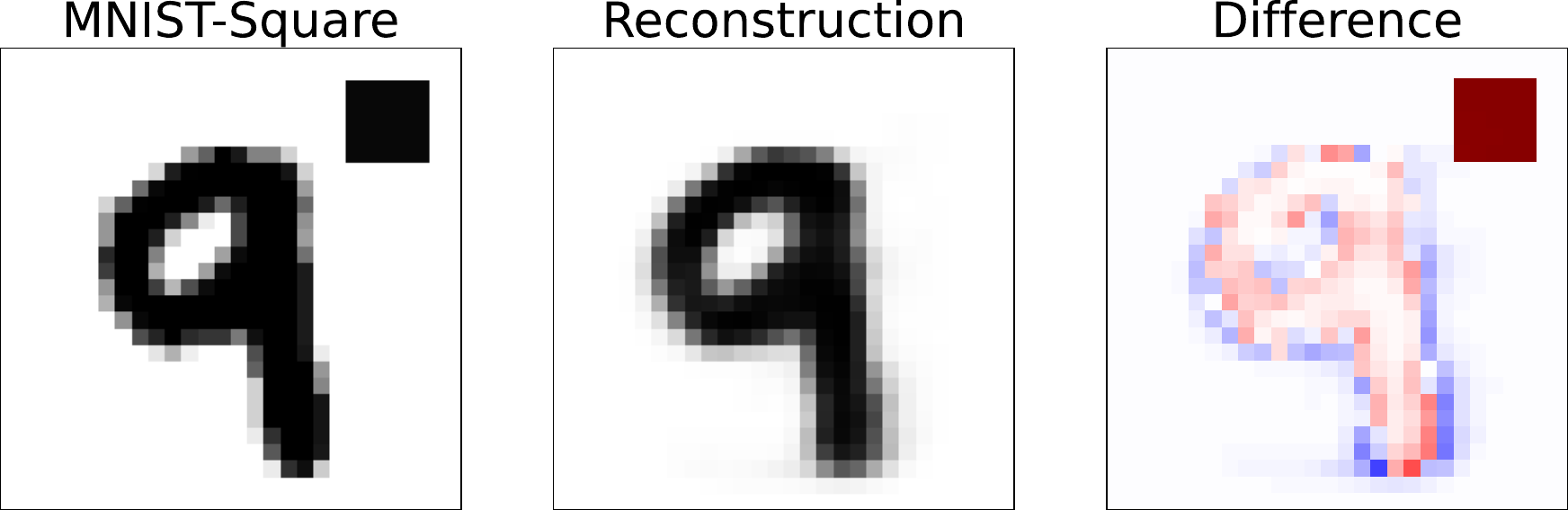}
        \label{subfig:mnist_attr11}
    }\hspace{0.1cm}
    \subfloat[\scriptsize MNIST-Gaussian ($\alpha = 0.7$)]{
        \includegraphics[width=0.3\textwidth]{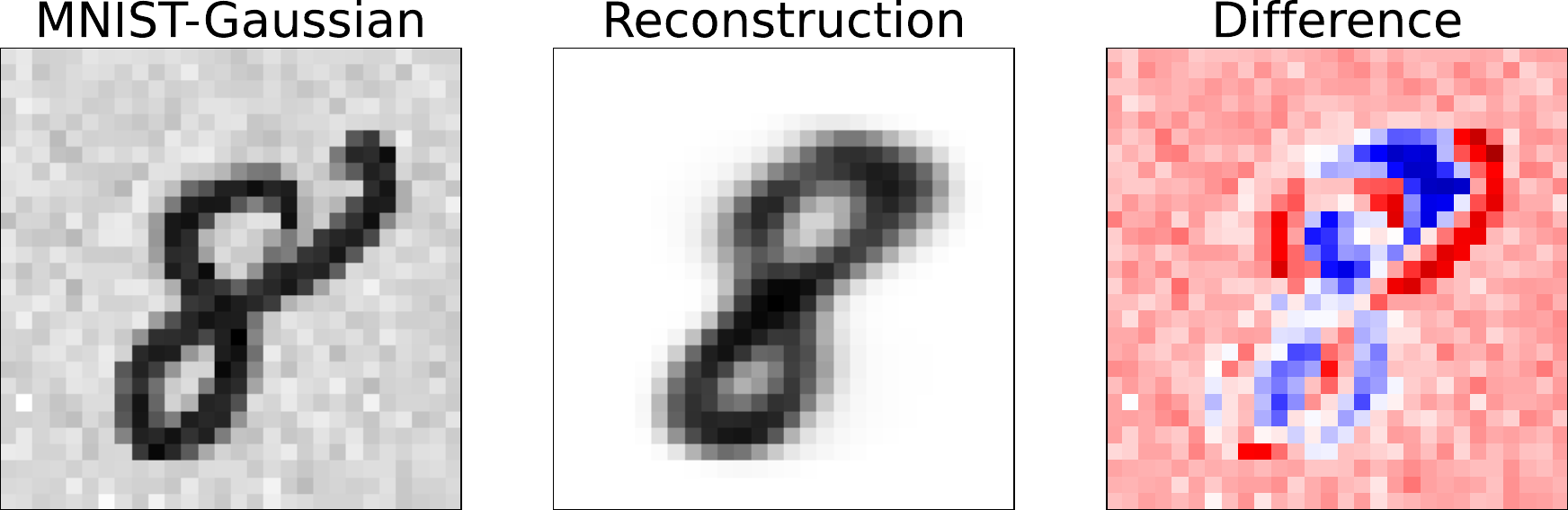}
        \label{subfig:mnist_attr12}
    }\hspace{0.1cm}
    \subfloat[\scriptsize MNIST-StripeOff ($\alpha = 11$)]{
        \includegraphics[width=0.3\textwidth]{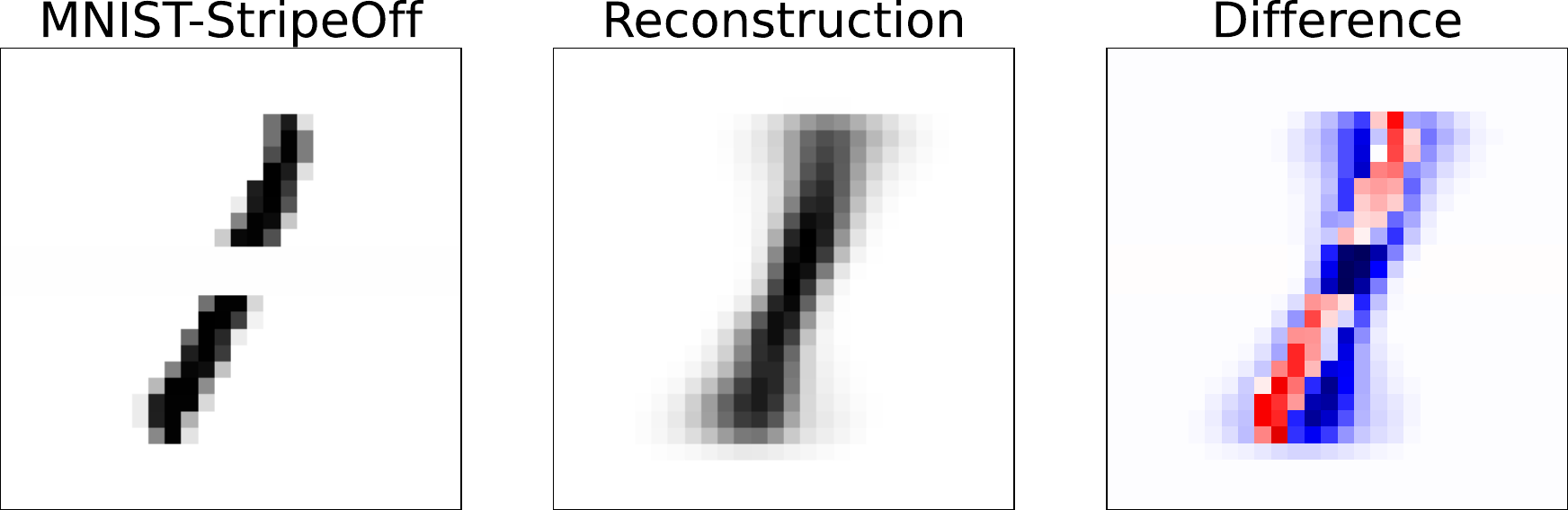}
        \label{subfig:mnist_attr13}
    }\\ \vspace{1mm}
    \subfloat[\scriptsize MNIST-StripeOn ($\alpha = 0.5$)]{
        \includegraphics[width=0.3\textwidth]{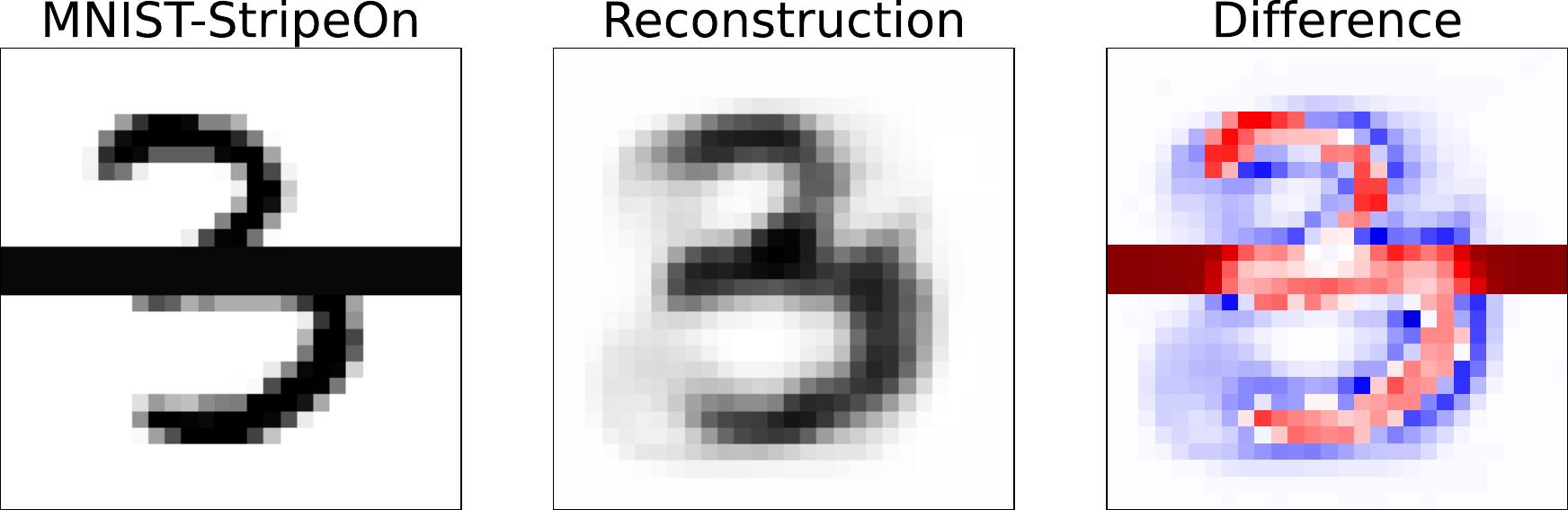}
        \label{subfig:mnist_attr14}
    }\hspace{0.1cm}
    \subfloat[\scriptsize KMNIST-Square ($\alpha = 0.1$)]{
        \includegraphics[width=0.3\textwidth]{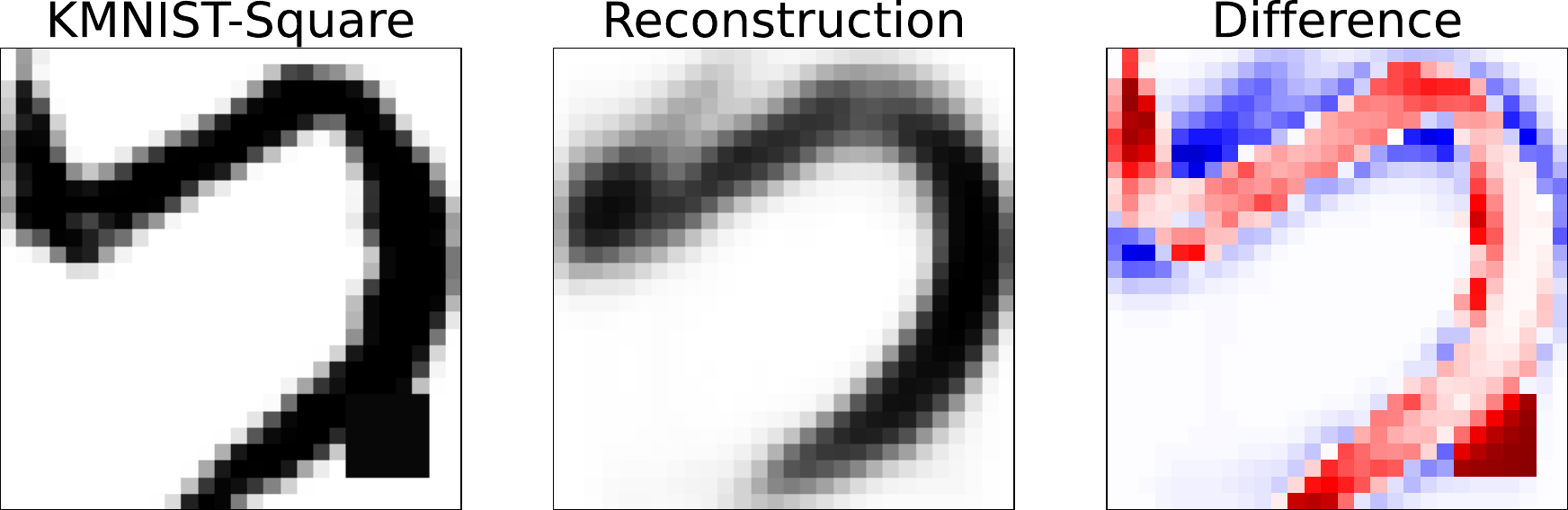}
        \label{subfig:kmnist_attr11}
    } \hspace{0.1cm}
    \subfloat[\scriptsize KMNIST-Gaussian ($\alpha = 0.7$)]{
        \includegraphics[width=0.3\textwidth]{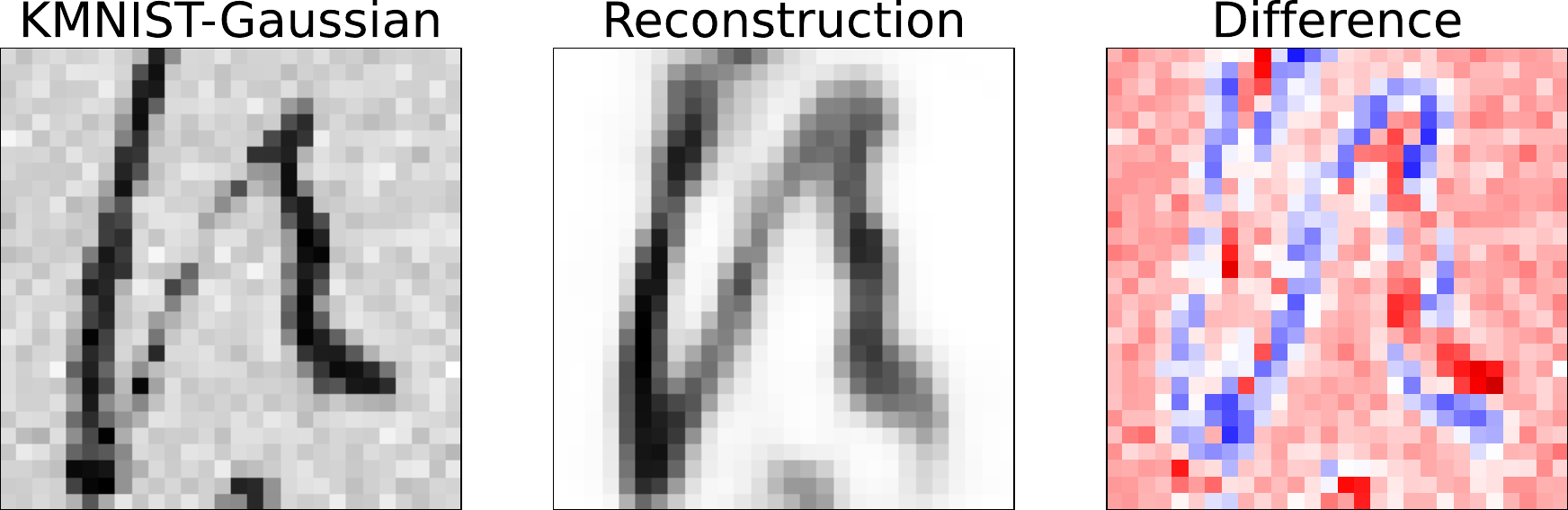}
        \label{subfig:kmnist_attr12}
    }\\ \vspace{1mm}
    \subfloat[\scriptsize KMNIST-StripeOn ($\alpha = 0.5$)]{
        \includegraphics[width=0.3\textwidth]{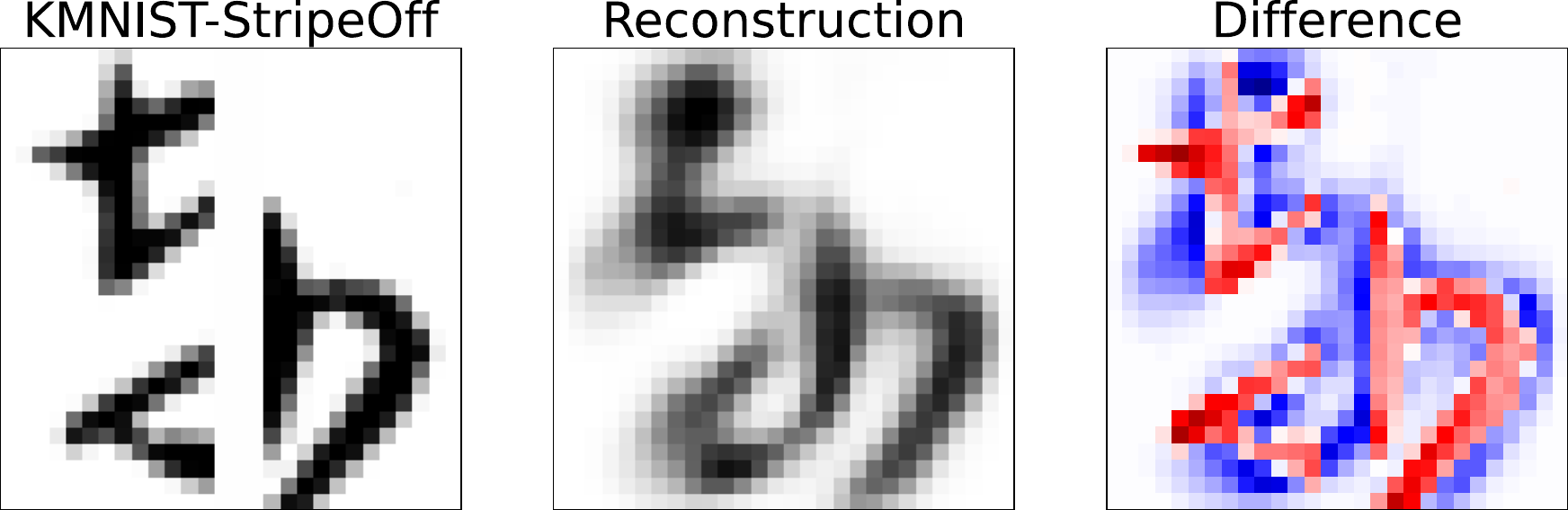}
        \label{subfig:kmnist_attr14}
    }\hspace{1.5cm}
    \subfloat[\scriptsize KMNIST-StripeOff ($\alpha = 11$)]{
        \includegraphics[width=0.3\textwidth]{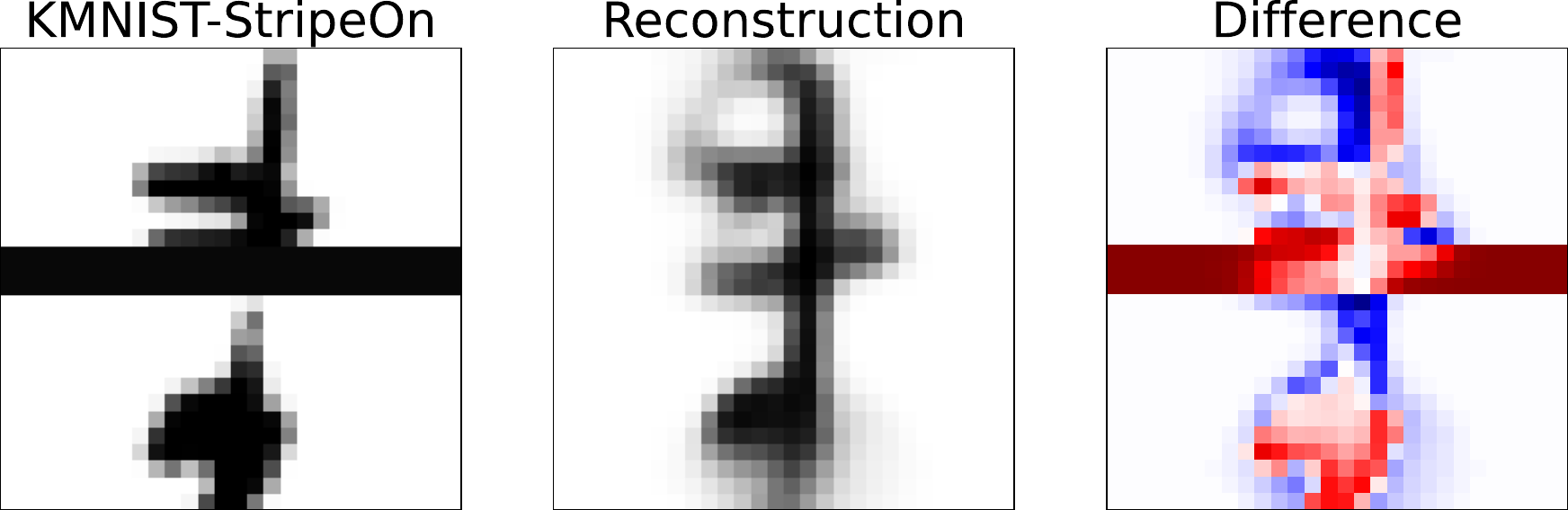}
        \label{subfig:kmnist_attr13}
    }    
    \caption{Batch of four visually modified samples from the MNIST and KMNIST datasets with their reconstructions and the corresponding attribution maps, featuring a distinct visual obstruction: a) \textit{Square}; b) \textit{Gaussian}; c ) \textit{StripeOff}; d) \textit{StripeOn};e) \textit{Square}; f) \textit{Gaussian}; g) \textit{StripeOff}; h) \textit{StripeOn}.}
    \label{fig:attr_example_kmnist}
\end{figure*}

To address the third research question and verify the quality of the attribution maps generated by our approach, as detailed in Algorithm \ref{algo:attribution}, we have employed a qualitative analysis of several visually modified samples from the MNIST and KMNIST datasets. Recalling from the approach detailed in Section \ref{sec:approach_attribution_technique}, our method provides an interpretability map by highlighting the regions of the image that push the original sample into the OoD space. Cold regions, represented by blue, indicate areas in the original input sample where certain features are missing compared to ID data. Conversely, hot regions, denoted by red, highlight areas in the original image that are not present in the ID data. The selected samples from the modified MNIST and KMNIST datasets are presented in Figure \ref{fig:attr_example_kmnist}. 

One initial result, consistent with the findings from \citep{ororbia2023predictive}, is that the latent space provides sufficient representation capabilities to offer accurate reconstructions without excessively losing the overall original shape. For instance, in Figure \ref{subfig:mnist_attr11}, the included square is clearly depicted as an artifact with the expected high temperature. Similarly, the horizontal stripe embedded in Figure \ref{subfig:mnist_attr14} is properly detected and erased in its respective reconstruction. Analogously, Figure \ref{subfig:mnist_attr13} showcases the algorithm's ability to detect missing regions, exposing the missing stripe using cold temperatures on the attribution map. Finally, in a mixed example from Figure \ref{subfig:mnist_attr12}, the algorithm detects the induced noise, providing a clean reconstruction of the sample. However, while the algorithm reports good quality attribution maps, all instances show slight distortions on the digit's borders. Nevertheless, the degree of deformation does not appear to affect the topology of the digit and is not significant enough for a human evaluator to misinterpret the artifacts with respect to the ID data. For instance, in its most extreme case, as observed in Figure \ref{subfig:mnist_attr14} with the \textit{StripeOn} modification, the artifact region remains the most noticeable area, easily distinguishable from the noisy contour of the number.

As pointed out in the training results and further confirmed in the OoD Detection results, the principal weakness of FFA becomes apparent in the KMNIST dataset, where the presence of complex features from the classes hinders the network's representation capabilities, impacting both accuracy and OoD detection results. As a consequence, a sharp drop in accuracy within the attribution maps is observed. This becomes evident when comparing results between the MNIST dataset and the KMNIST dataset, where the shape of KMNIST reconstructions is less accurate than those obtained from the MNIST dataset, resulting in noisier attribution maps. Nevertheless, the overall shape of the figure is still consistently reconstructed in most of the samples that are correctly classified.

One such instance is observed in Figure \ref{subfig:kmnist_attr14}, where the upper and lower limits of the character have been distorted, while the stripe artifact is properly removed and highlighted. A similar effect is present in Figure \ref{subfig:kmnist_attr11}, where our algorithm can properly detect the embedded square, showing the highest temperature, while the resulting character outline has been slightly modified but remains within the original class distribution. In contrast, a greater shift can be observed when the modifications to the image are more drastic. Such is the case for Figure \ref{subfig:kmnist_attr13}, where the obtained reconstruction is able to fill the missing stripe, but the contour appears blurred and several previously sharp features are missing, such as the upper stroke. Finally, in Figure \ref{subfig:kmnist_attr12}, the algorithm is able to remove a significant portion of the noise with minimal changes to the shape of the characters and only a small amount of blur. Although all samples exhibit slight perturbations and do not present clean attribution, the discriminative regions of the embedded artifacts can be clearly detected. Therefore, our results provide a positive answer to RQ3, showcasing the capability of our algorithm to create interpretable attribution maps that explain the features pushing the samples into the OoD regions. 

\subsection{Limitations observed in the experiments}
\label{sec:limitations}
FFA shows promise in enabling forward-only learning, particularly in datasets with high inter-class distance. This is evident in its strong performance on datasets such as MNIST, KMNIST, and EMNIST. However, when applied to image-based data using convolutional architectures, early experiments suggest that the algorithm struggles to converge to optimal solutions. It often either significantly underperforms or fixates on predicting a single class. This limitation makes applying FFA to natural image datasets like CIFAR-10 particularly challenging. In our work, we investigate the use of ResNet18 as a backbone, incorporating FFA as a secondary projection network to reshape the latent space into a more manageable geometry. Our results demonstrate that FFA can still effectively distinguish between ID and OoD data, even when the backbone is not explicitly trained for a specific task but instead derived from a pretrained general image dataset. However, the need for additional networks partially undermines the method’s independence, underscoring the necessity for further research into alternative implementations of FFA to enhance its viability within convolutional architectures.

\section{Conclusions}\label{sec:conc}

This work introduces FF-OoD, a novel Out-of-Distribution (OoD) detection method for Spiking Neural Networks (SNNs) that employs the characteristic topology emerging from networks trained using the Forward-Forward Algorithm (FFA). To achieve this, we extend the original formulation of FFA's goodness functions to accommodate the additional temporal dimension inherent to SNNs and study the novel cluster structure that arises among the so-called negative data. Given the significant separation observed between distinct classes within the latent space of these networks, FF-OoD computes the distance between input samples and precomputed representative manifolds of known classes. In addition, we provide theoretical insights into how high intra-class variability affects the accuracy of distance-based scoring, reversing the scores of ID and OoD samples, and offer a strategy to overcome this issue. To enhance the explainability and trustworthiness of our method, we also propose a novel attribution technique that explores the latent space to generate attribution maps, highlighting features that deviate from those seen in In-Distribution (ID) classes. To assess the competitiveness of our FFA-based approach, we trained networks on multiple well-known datasets (e.g., KMNIST, EMNIST, CIFAR-10) and evaluated FF-OoD using OoD samples from conventional baselines (e.g., Omniglot, NotMNIST). Additionally, we conducted a qualitative analysis of our interpretability method by manually introducing artifacts into training samples and evaluating the quality of the resulting attribution maps. Our results demonstrate that the latent space of networks trained with FFA naturally separates OoD instances from ID class manifolds, enabling effective OoD detection in spiking neural networks. Furthermore, we showcase the effectiveness of our attribution method in accurately highlighting anomalies injected into input images, providing an intuitive and human-interpretable way to identify the regions responsible for pushing a sample outside its class distribution.

\paragraph{Limitations and Future Work} To obtain a faithful representation of the latent space, our method requires a large number of latent representatives from each class, making it memory-intensive when dealing with datasets with many classes. Our primary focus is on developing mechanisms to reduce intra-class distance, thereby creating a more tractable geometry, especially for instances with small ID datasets. Another key direction for future work stems from the selectivity observed in these networks, which could be leveraged for lifelong learning tasks. We aim to develop a detection mechanism to identify when novel clusters of OoD data emerge and to label them as novel classes.

\section*{Acknowledgements} The authors thank the Basque Government for its funding support via the consolidated research groups MATHMODE (ref. T1256-22) and D4K (ref. IT1528-22), and the BEREZ-IA ELKARTEK project (ref. KK-2023/00012). Erik B. Terres-Escudero acknowledges the funding support from TECNALIA Research \& Innovation, provided to employees pursuing their doctoral thesis.

{\appendix
\section{Reversal of Order in FF-OoD} \label{ap:proof_equation}

This appendix formalizes the mathematical arguments that drive the reversal of the order in the scoring function. As pointed out in Section \ref{sec:approach_ff_scp}, the main limitation of the initial scoring function depicted in Equation \eqref{eq:initial_ood_cond} arises when, given a randomly selected sample from the ID dataset \(X\), the following inequality holds:
\begin{equation}
\min_{c \in C} \widehat{d}(\textbf{0}, L_{c,c} ) \le \mathbb{E}\left[\min_{c \in C} \widehat{d}(\mathcal{N}_1(X, c), L_{c,c} )\right].
\end{equation}

For instance, this reversal effect of the order predominantly occurs in OoD samples that were identified as negative inputs by the network, resulting in a near-zero latent state. Formally, let \(\mathcal{N}_k(\cdot, \cdot)\) be an \textit{low-normed negative network}, defined as a network in which the following expression holds:
\begin{equation}
\label{eq:hold_condition_aen}
    \mathbb{E}\Big[ \| \mathcal{N}_1(\bm{X}_{\ominus}, c) \|_2 \Big] < \epsilon,
\end{equation}
where \(\bm{X}_{\ominus}\) denotes a random negative sample and \(\epsilon\) represents a sufficiently low value.

When networks hold this property, we can verify that the discrete set of latent representatives \(\mathcal{L}_{c,p}\), given two distinct \(c,p \in \mathcal{C}\), will have near-zero mean norm after the initial filtering step of the FF-OoD algorithm. This effect occurs as the original set of negative representatives, which is predominantly composed of elements with a norm less than \(\epsilon\), will have all high-normed elements removed after the initial filtering step. Therefore, we can assert that \(\mathcal{L}_{c,p} \sim \textbf{0}\) whenever $c \not = p$.

For notational purposes, let $\mathcal{N}_1(x_{\ominus}, p)$ be denoted as $\bm{\ell}^p_{\ominus}$, and $\mathcal{N}_1(x_{\oplus}, p)$ as $\bm{\ell}^p_{\oplus}$ . Then, given a \textit{low-normed negative network}, the scoring function can be simplified to:
\begin{multline}
s(x_{\ominus}) = \min_{p \in C} \sum_{c \in C} \widehat{d}(\bm{\ell}^p_{\ominus}, \mathcal{L}_{c,p} ) \approx \min_{p \in C} \sum_{c \in C} \widehat{d}(\textbf{0}, \mathcal{L}_{c,p} ).
\end{multline}

Additionally, since $\widehat{d}(\textbf{0}, \mathcal{L}_{c,p}) \sim \widehat{d}(\textbf{0},\textbf{0}) = 0$ when $p\not=c$, the expression can be further simplified to:
\begin{equation}
s(x_{\ominus}) = \min_{p \in C} \sum_{c \in C} \widehat{d}(\textbf{0}, \mathcal{L}_{c,p} ) \approx \min_{p \in C}  \widehat{d}(\textbf{0}, \mathcal{L}_{p,p} ).
\end{equation}

This result implies that the score \(s(x_{\ominus})\)  of an OoD sample \(x_{\ominus}\) with near-zero norm will be almost equal to the distance between the origin of the space to the closest point among the positive latent clusters \(\mathcal{L}_{p,p}\), with \(p \in \mathcal{C}\).

Similarly, given the same network $\mathcal{N}$ and an ID sample $x_{\oplus}$, the scoring function $s(x_{\oplus})$ can be simplified for negative samples. Using Equation \eqref{eq:hold_condition_aen} and applying the same logic as with the OoD samples, the summation over the different latent clusters can be reduced to only the value of the latent cluster forwarded to the real label, i.e.:
\begin{equation}
\min_{p \in C} \sum_{c \in C} \widehat{d}(\bm{\ell}^p_{\oplus}, \mathcal{L}_{c,p} ) \approx 
\min_{p \in C} \widehat{d}(\bm{\ell}^p_{\oplus}, \mathcal{L}_{p,p} ),
\end{equation}
which implies that $s(x_{\oplus})$ is approximated by the minimal distance between the latent vector of $x_{\oplus}$ and the closest latent point in $L_p^p$. If the distance between points inside the cluster is sufficiently large, to the extent where it is surpasses the value $\min_{p \in C}  \widehat{d}(\textbf{0}, \mathcal{L}_{p,p} )$, the value of $s(x_{\oplus})$ will be greater than $s(\ominus)$, thereby showing the reversing effect in the order of the OoD scores.

\section{OoD Detection Results without Order Reversal Considerations}
\label{ap:ood_different_sx}

This appendix presents the results obtained when employing the FF-OoD algorithm when employing the $s(\cdot)$ scoring function. As previously mention in Section \ref{sec:approach_ff_scp}, the function $s(\cdot)$ does not properly measure the distance in scenarios where the intra-cluster distance is greater than the distance between the zero vector and each sample of positive latent clusters, which is formally proven in \ref{ap:proof_equation}. As expected, the results from Table \ref{tab:snn_ood_results_comp_s_x} evidence the reduced OoD detection accuracy when compared to the ones from Table \ref{tab:snn_ood_results_comp_final}, which relies on the refined scoring function $S(\cdot)$.

\begin{table*}[t]
    \definecolor{mygray}{RGB}{80,80,80}
    \newcommand{\ceco}{\cellcolor{blue!6}}
    \newcommand{\teco}[1]{\textbf{\textcolor{mygray}{#1}}}
    \centering

    \caption{Results of the FF-OoD algorithm using the $s(\cdot)$ of our three proposed networks: the Analog FF-Network; the Unbounded FF-Network, denoted as $G_\infty(\cdot)$; and the Bounded FF-Network, denoted as $G_0(\cdot)$. Data from the SCP and ODIN has been extracted from the paper \citep{seras2022novel}. The ANN results are only shown as a baseline to compare ANN vs SNN.}    
    \label{tab:snn_ood_results_comp_s_x}

    \resizebox{\textwidth}{!}{
    \begin{tabular}{clcccccp{0.2cm}cccccp{0.2cm}ccccc}
    \toprule[2pt]
        \midrule
        & & \multicolumn{5}{c}{AUROC $\uparrow$}  & & \multicolumn{5}{c}{ AUPR $\uparrow$}    & & \multicolumn{5}{c}{FPR95 $\downarrow$}      \\ \cmidrule{3-7} \cmidrule{9-13} \cmidrule{15-19} 
        \makecell[c]{ID\\Dataset} & \makecell[c]{OOD\\Dataset} &    \thead{FF-OoD \\ {\small Analog}} &  \thead{FF-OoD \\ {\small $G_{\infty}(\cdot)$}} &  \thead{FF-OoD \\ {\small $G_{0}(\cdot)$}}  & SCP & ODIN   & &  \thead{FF-OoD \\ {\small Analog}} &  \thead{FF-OoD \\ {\small $G_{\infty}(\cdot)$}} &  \thead{FF-OoD \\ {\small $G_{0}(\cdot)$}} & SCP & ODIN  & & \thead{FF-OoD \\ {\small Analog}} & \thead{FF-OoD \\ {\small $G_{\infty}(\cdot)$}} & \makecell[c]{FF-OoD \\ {\small $G_{0}(\cdot)$}} & SCP & ODIN \\  
        \midrule
         \multirow{5}{*}{MNIST}
         & Omniglot   & \teco{83.19} & 90.25 & 88.80& 71.56  & \textbf{94.72} & & \teco{89.24} & 92.19 & 91.67& 71.90 & \textbf{94.28} & & \teco{50.28} & 29.65 & 36.03& 88.93 & \textbf{29.08} \\
         & notMNIST   & \teco{\textbf{99.99}} & \textbf{99.98} & 99.98& 99.14  & 64.45 & & \teco{99.99} & 99.99 & 99.99& 99.01 & 57.16 & & \teco{\textbf{0.00}} & 0.00 & 0.00& 03.60 & 86.65 \\
         & FMNIST     & \teco{93.04} & 97.88 & \textbf{99.37}& 91.21  & 73.97 & & \teco{94.62} & 97.73 & \textbf{99.40}& 87.15 & 65.29 & & \teco{50.28} & 11.22 & \textbf{3.38}& 24.95 & 77.11 \\
         & KMNIST     & \teco{97.87} & 98.39 & \textbf{99.40}& 94.89  & 73.97 & & \teco{98.12} & 98.38 & \textbf{99.42}& 94.92 & 76.27 & & \teco{13.49} & 8.92 & \textbf{3.39}& 24.99 & 72.87   \\
         & Letters    & \teco{84.80} & 91.26 & \textbf{96.88}& 88.11  & 81.43 & & \teco{92.46} & 95.55 & \textbf{98.48}& 79.49 & 64.19 & & \teco{61.61} & 42.22 & \textbf{15.25}& 51.66 & 73.89 \\
        \midrule
        \multirow{5}{*}{FMNIST}
        & MNIST       & \teco{99.07} & \textbf{99.03} & 98.59& 92.60  & 95.98 & & \teco{98.72} & \textbf{98.42} & 97.99& 93.18 & 95.74 & & \teco{3.65} & \textbf{3.27} & 5.04& 35.48 & 17.37   \\
         & Omniglot   & \teco{98.83} & \textbf{98.72} & 95.46& 84.75  & 97.86 & & \teco{99.09} & \textbf{98.78} & 96.76& 85.37 & 97.83 & & \teco{4.72} & \textbf{4.09} & 21.08& 63.55 & 09.97 \\
         & notMNIST   & \teco{99.85} & \textbf{99.93} & 99.92& 98.87  & 61.70 & & \teco{99.93} & \textbf{99.96} & 99.96& 99.14 & 61.75 & & \teco{0.10} & \textbf{0.07} & 0.27& 02.47 & 92.33 \\
         & KMNIST     & \teco{97.27} & 97.83 & \textbf{98.10}& 84.56  & 88.36 & & \teco{97.13} & 97.48 & \textbf{97.86}& 82.41 & 88.84 & & \teco{13.34} & 9.46 & \textbf{8.75}& 53.10 & 57.39 \\
         & Letters    & \teco{97.80} & \textbf{98.32} & 97.11& 92.83  & 91.83 & & \teco{98.72} & \textbf{98.84} & 98.51& 88.10 & 85.15 & & \teco{10.13} & \textbf{7.05} & 16.06& 34.49 & 39.82 \\
        \midrule
        \multirow{5}{*}{KMNIST}
         & MNIST      & \teco{87.35} & \textbf{87.36} & 61.13& 61.85 & 85.79 & & \teco{80.92} & 82.12 & 53.79& 61.64 & \textbf{88.57} & & \teco{36.55} & \textbf{41.84} & 75.55& 89.34 & 51.64   \\
         & Omniglot   & \teco{85.68} & 82.28 & 54.87& 58.54 & \textbf{96.43} & & \teco{88.94} & 85.73 & 62.51& 59.68 & \textbf{97.25} & & \teco{47.07} & 51.68 & 69.27& 92.58 & \textbf{08.09} \\
         & notMNIST   & \teco{92.79} & 95.74 & \textbf{99.08}& 96.88 & 67.24 & & \teco{96.88} & 98.04 & \textbf{99.50}& 96.68 & 67.59 & & \teco{60.16} & 34.05 & \textbf{3.94}& 09.52 & 81.37 \\
         & FMNIST     & \teco{17.92} & 32.95 & 66.42& 75.00 & \textbf{76.73} & & \teco{33.85} & 39.38 & 59.32& 72.85 & \textbf{78.58} & & \teco{97.96} & 96.78 & 72.01& \textbf{66.14} & 70.67 \\
         & Letters    & \teco{71.46} & 80.69 & 81.66& 78.34 & \textbf{85.14} & & \teco{79.16} & \textbf{85.23} & 83.94& 64.47 & 80.65 & & \teco{65.63} & 51.44 & \textbf{46.22}& 71.49 & 53.51 \\
        \midrule
        \multirow{5}{*}{Letters}
        & MNIST       & \teco{87.64} & 83.26 & \textbf{83.65}& 69.06 & 80.98 & & \teco{80.61} & 76.33 & 71.52& 82.30 & \textbf{88.73} & & \teco{53.65} & 71.52 & 68.37& 88.83 & \textbf{68.13}   \\
         & Omniglot   & \teco{79.73} & 81.14 & 85.28& 77.60 & \textbf{97.11} & & \teco{77.35} & 76.11 & 77.66& 88.04 & \textbf{98.54} & & \teco{68.44} & 59.40 & 40.72& 87.11 & \textbf{15.54} \\
         & notMNIST   & \teco{99.98} & \textbf{\textbf{99.99}} & 99.95& 98.63 & 58.08 & & \teco{99.98} & 99.99 & 99.95& 98.90 & 69.65 & & \teco{\textbf{0.00}} & 0.00 & 0.00& 07.48 & 93.94 \\
         & FMNIST     & \teco{90.09} & \textbf{95.98} & 91.01& 86.33 & 71.30 & & \teco{88.94} & \textbf{94.88} & 84.47& 92.08 & 83.66 & & \teco{70.62} & \textbf{27.30} & 35.04& 47.30 & 88.04\\
         & KMNIST     & \teco{99.21} & \textbf{99.65} & 97.27& 86.31 & 77.60 & & \teco{98.63} & \textbf{99.35} & 95.21& 92.15 & 85.47 & & \teco{3.83} & \textbf{1.42} & 14.93& 51.41 & 77.88 \\
        \midrule
         &  \textbf{Average} & \teco{88.18} & \textbf{90.53} & 89.70 & 84.35 & 81.03 & & \teco{89.66} & \textbf{90.72} & 88.40 & 84.47 & 81.25 & & \teco{35.58} & 27.57 & \textbf{26.76}& 49.72 & 58.26 \\
         &  \textbf{Median} & \teco{92.91} & 96.90 & \textbf{97.00} & 86.32 & 81.20 & & \teco{95.75} & \textbf{97.61} & 97.31 & 87.59 & 84.40 & & \teco{41.81} & 19.26 & \textbf{15.65}& 51.53 & 69.4 \\
        \bottomrule
        \end{tabular}}    
\end{table*}

\section{Latent State of FFA} \label{ap:latent_geometry}

To analyze the latent space of models trained using FFA, we rely on the trained models used for the OoD detection experiments, specifically the analog model using ReLU activation functions and the bounded goodness SNN. These two models have been selected as they encompass both activity regimes observed during the experimentation: the sparse latent space and the bounded latent space. To avoid redundancy in the latent space, and given the similar latent behavior arising from the different datasets, we employ random batches of data from the MNIST dataset to extract the latent spaces.

The plot showing the latent vectors obtained from the analog ReLU network is depicted in Figure \ref{fig:ann_latent}. Consistent with the findings of \citet{tosato2023emergent} and \citet{yang2023theory}, the latent vectors emerging from these networks are characterized by sparse activity regimes and high neural specialization. As observed in the images, most latent vectors are composed of only a small subset of active neurons, which represent key distinctive features of the different classes. For instance, given that the latent vectors of the images are ordered by class, it can be observed that the set of neurons that activate when presented with each input distribution is highly consistent among the batch. Similarly, each neuron is usually only active when presented with a small subset of the available input classes, often being active with only one. 

Analogously, Figure \ref{fig:snn_latent} depicts the latent activity from the aforementioned SNN. This image showcases denser latent vectors, with a larger amount of active neurons per instance, still keeping the class-based activation patterns within the activity. Upon early experimentation, the emergence of these less sparse representations appears to be a byproduct of the bounded regimes of the goodness functions. Due to the inability to achieve large goodness values through a small set of unbound neurons, the network employs on a large set of neurons that ends up adding to higher goodness scores. Furthermore, bounded models create non-dormant negative samples, which highly contrast with the near-zero latent activity from negative samples in unbounded models.  While the results in Table \ref{tab:Accuracy_ANN_SNN} indicate a small decrease in the accuracy from these networks, the increased bound of activity yields positive outcomes in terms of representability, as shown in the OoD detection results in Table \ref{tab:snn_ood_results_comp_final}.

\begin{figure}[h]
    \centering
    \subfloat[Analog Network.]{
        \includegraphics[width=0.47\columnwidth]{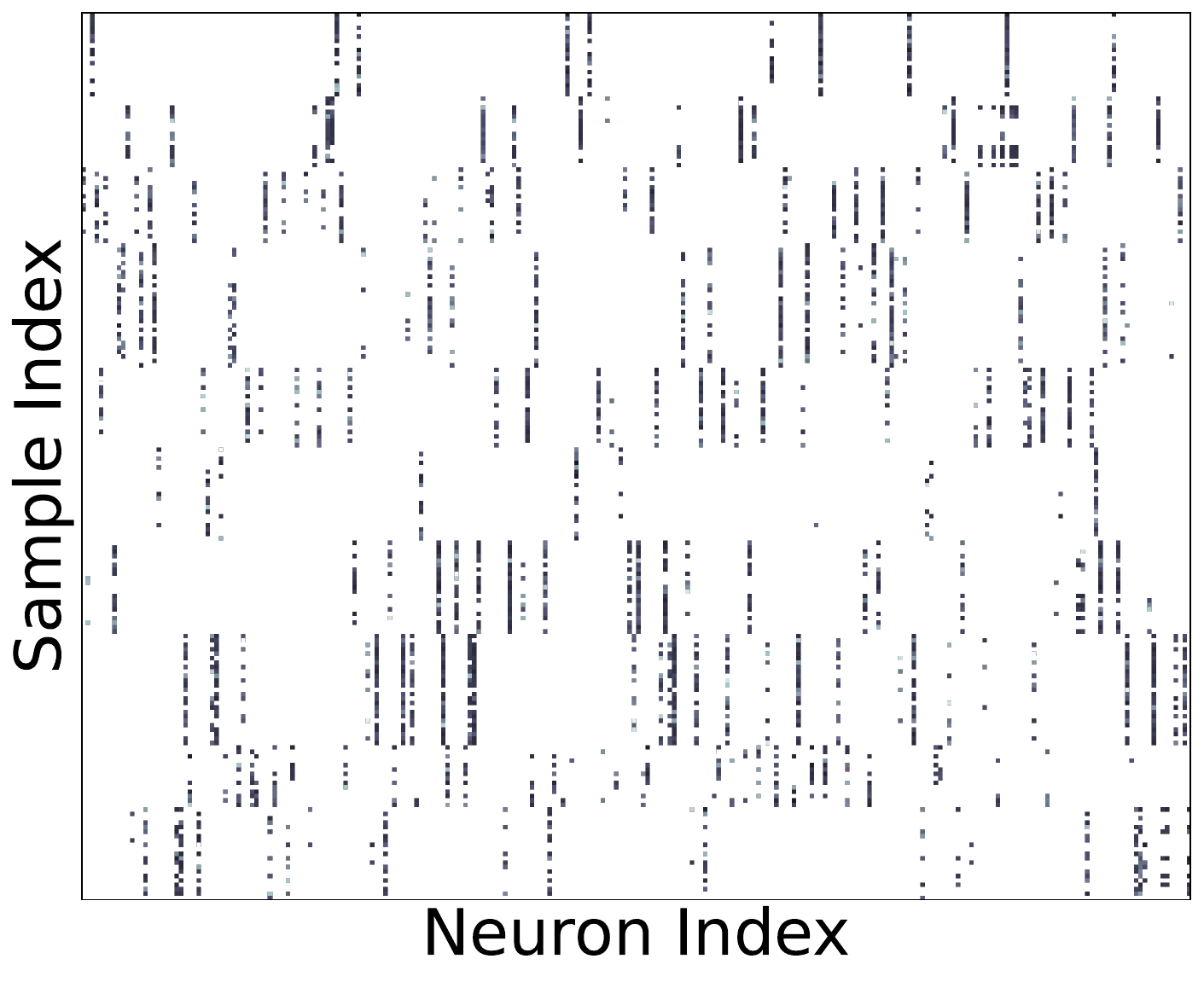}
        \label{fig:ann_latent}
    }
    \subfloat[Spiking Network.]{
        \includegraphics[width=0.47\columnwidth]{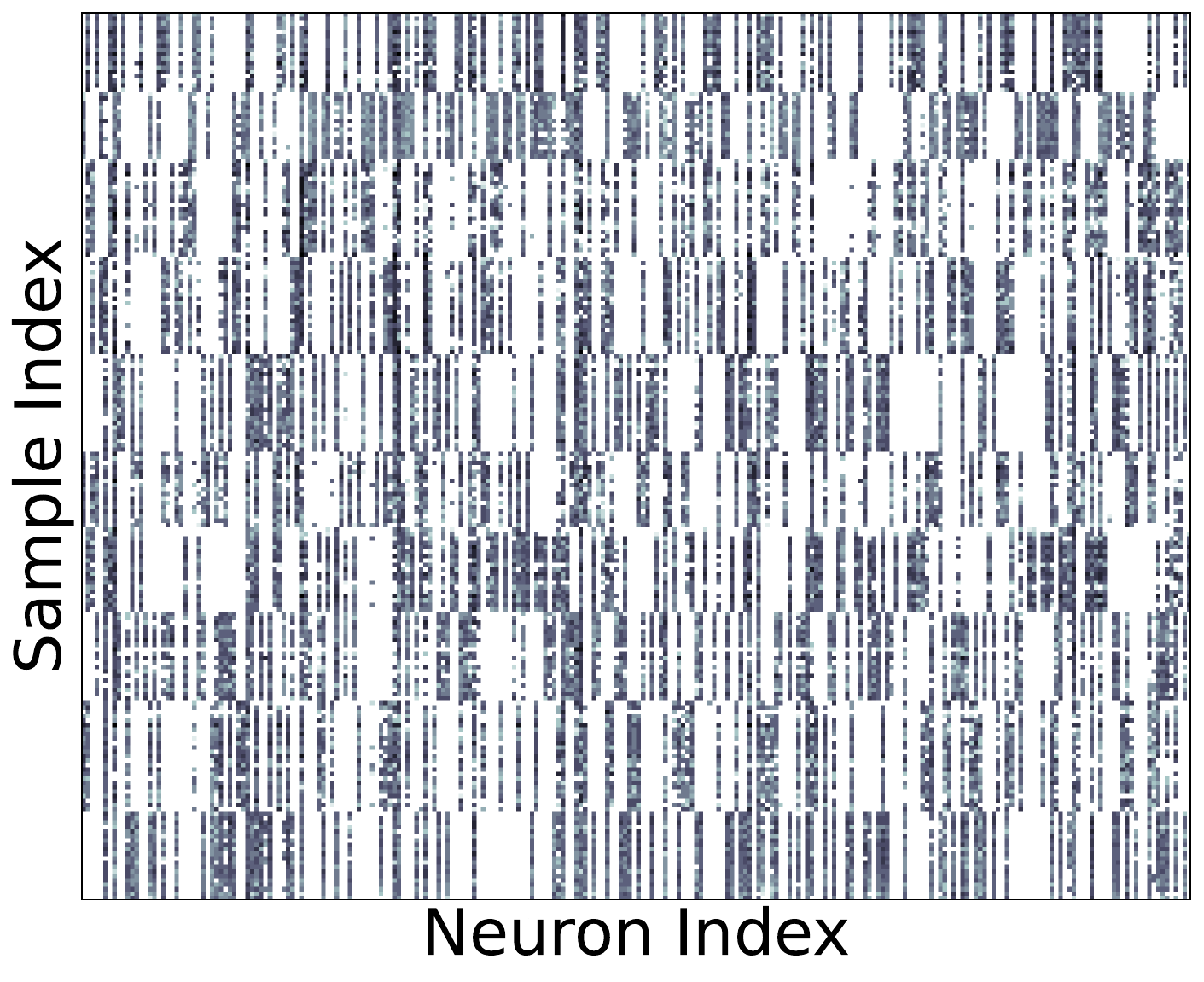}
        \label{fig:snn_latent}
    }
    \captionsetup{font=small, skip=7pt}  
    \caption{Latent vectors of the first layer of 256 random samples ordered by class extracted from networks trained with FFA. Only the first 250 neurons are displayed.}
    \label{fig:ann_latent_max}
\end{figure}

To provide a more in-depth analysis of the geometrical properties of these latent vectors, we employed dimensionality reduction methods to present approximate depictions of the structure of the latent space. We relied on the T-SNE algorithm \citep{van2008visualizing}, using 1024 randomly chosen samples from MNIST, with latent vectors obtained by using all 10 available class labels. This approach allows us to explore the geometrical relationship between positive and negative samples, beyond the activity differences.

The latent space of the analog and the spiking networks are depicted in Figure \ref{subfig:ann_tsne_mini} and Figure \ref{subfig:snn_tsne_mini}, respectively. Once again, our results align with those observed by other researchers: positive samples cluster together, farther from the zero vector, and are highly separated from negative samples. Similarly, the position of the negative set also appears to contain a clearly defined structure, with most samples staying close to the origin. In the spiking case, one clear property that arises as a consequence of the denser latent activity is the clustering of negative latent vectors into additional subclusters, which are formed by samples belonging to pairs of real classes and embedded labels. This latent geometry offers more precise results when measuring distances to OoD samples, as the clusters of class/label representatives provide closer distances among samples from the same distribution. While this effect increases the separation between positive and negative samples, this geometric property does not necessarily translate to enhanced accuracy, as the denser latent vectors also reduces the specialization of neurons in the layer.

\begin{figure}[h]
    \centering
    \subfloat[Analog Network.]{
        \includegraphics[width=0.47\columnwidth]{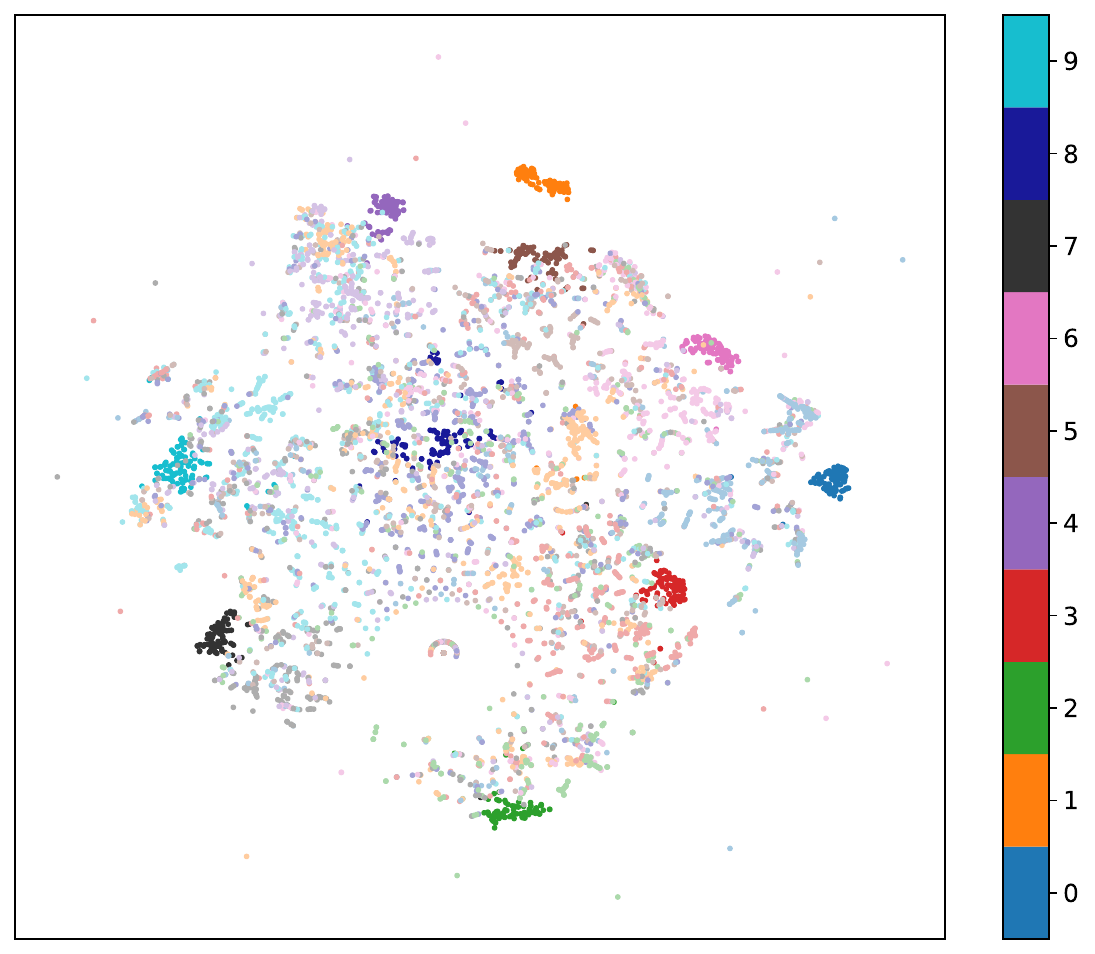}
        \label{subfig:ann_tsne_mini}
    }
    \subfloat[Spiking Network.]{
        \includegraphics[width=0.47\columnwidth]{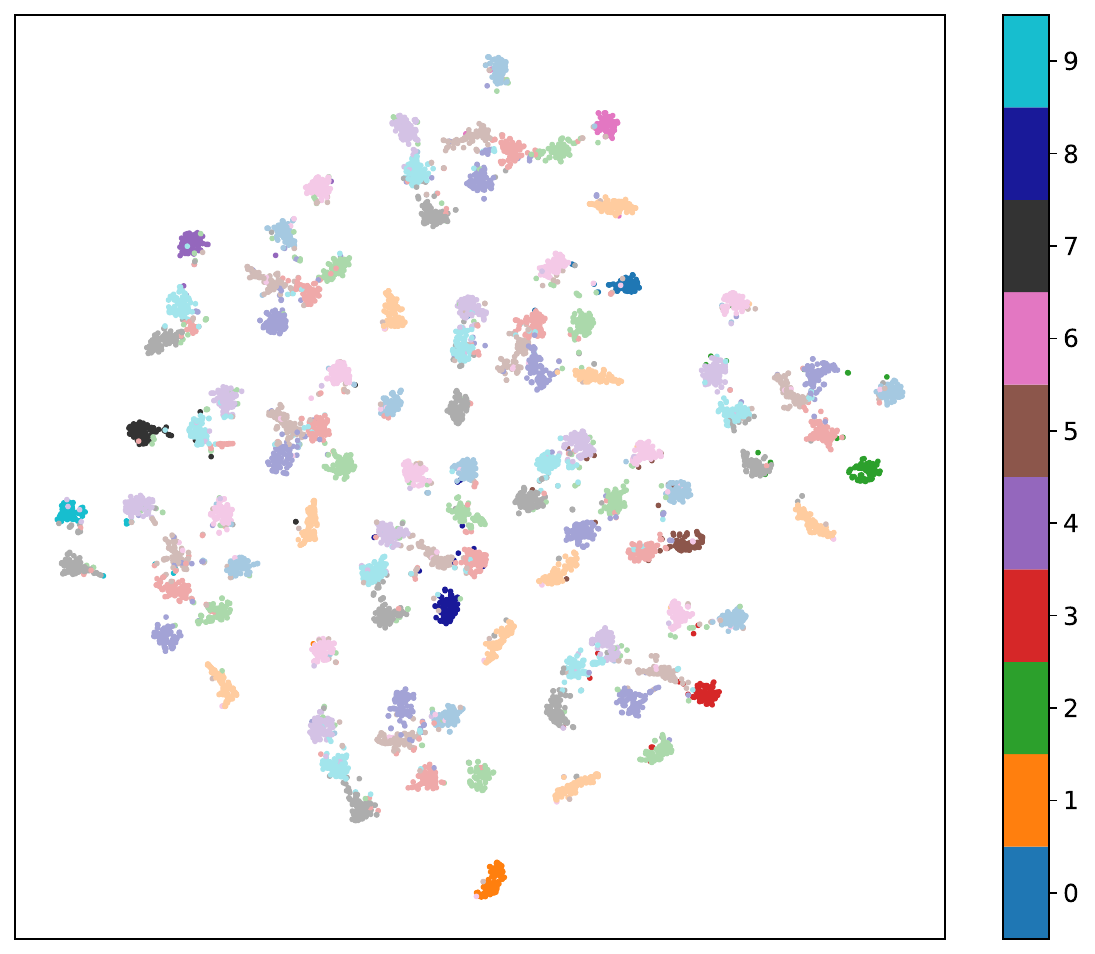}
        \label{subfig:snn_tsne_mini}
    }
    \captionsetup{font=small, skip=7pt}  
    \caption{T-SNE projection of the 10240 latent vectors extracted from the first layer of the analog neural network over the MNIST dataset. Positive vectors are illustrated using darker colors to differentiate them from negative samples.}
    \label{fig:ann_tsne_max}
\end{figure}

}

\bibliographystyle{elsarticle-harv} 
\bibliography{references} 

\begin{thebibliography}{42}
\expandafter\ifx\csname natexlab\endcsname\relax\def\natexlab#1{#1}\fi
\providecommand{\url}[1]{\texttt{#1}}
\providecommand{\href}[2]{#2}
\providecommand{\path}[1]{#1}
\providecommand{\DOIprefix}{doi:}
\providecommand{\ArXivprefix}{arXiv:}
\providecommand{\URLprefix}{URL: }
\providecommand{\Pubmedprefix}{pmid:}
\providecommand{\doi}[1]{\href{http://dx.doi.org/#1}{\path{#1}}}
\providecommand{\Pubmed}[1]{\href{pmid:#1}{\path{#1}}}
\providecommand{\bibinfo}[2]{#2}
\ifx\xfnm\relax \def\xfnm[#1]{\unskip,\space#1}\fi
\bibitem[{Abbott(1999)}]{abbott1999lapicque}
\bibinfo{author}{Abbott, L.F.}, \bibinfo{year}{1999}.
\newblock \bibinfo{title}{Lapicque’s introduction of the integrate-and-fire model neuron (1907)}.
\newblock \bibinfo{journal}{Brain Research Bulletin} \bibinfo{volume}{50}, \bibinfo{pages}{303--304}.
\bibitem[{Arrieta et~al.(2020)Arrieta, D{\'\i}az-Rodr{\'\i}guez, Del~Ser, Bennetot, Tabik, Barbado, Garc{\'\i}a, Gil-L{\'o}pez, Molina, Benjamins, Chatila and Herrera}]{arrieta2020explainable}
\bibinfo{author}{Arrieta, A.B.}, \bibinfo{author}{D{\'\i}az-Rodr{\'\i}guez, N.}, \bibinfo{author}{Del~Ser, J.}, \bibinfo{author}{Bennetot, A.}, \bibinfo{author}{Tabik, S.}, \bibinfo{author}{Barbado, A.}, \bibinfo{author}{Garc{\'\i}a, S.}, \bibinfo{author}{Gil-L{\'o}pez, S.}, \bibinfo{author}{Molina, D.}, \bibinfo{author}{Benjamins, R.}, \bibinfo{author}{Chatila, R.}, \bibinfo{author}{Herrera, F.}, \bibinfo{year}{2020}.
\newblock \bibinfo{title}{Explainable artificial intelligence ({XAI}): Concepts, taxonomies, opportunities and challenges toward responsible {AI}}.
\newblock \bibinfo{journal}{Information Fusion} \bibinfo{volume}{58}, \bibinfo{pages}{82--115}.
\bibitem[{Bergman et~al.(2020)Bergman, Cohen and Hoshen}]{bergman2020deep}
\bibinfo{author}{Bergman, L.}, \bibinfo{author}{Cohen, N.}, \bibinfo{author}{Hoshen, Y.}, \bibinfo{year}{2020}.
\newblock \bibinfo{title}{Deep nearest neighbor anomaly detection}.
\newblock \bibinfo{journal}{arXiv:2002.10445} .
\bibitem[{Bitar et~al.(2023)Bitar, Rosales and Paulitsch}]{bitar2023gradient}
\bibinfo{author}{Bitar, A.}, \bibinfo{author}{Rosales, R.}, \bibinfo{author}{Paulitsch, M.}, \bibinfo{year}{2023}.
\newblock \bibinfo{title}{Gradient-based feature-attribution explainability methods for spiking neural networks}.
\newblock \bibinfo{journal}{Frontiers in Neuroscience} \bibinfo{volume}{17}.
\bibitem[{Bossard et~al.(2014)Bossard, Guillaumin and Van~Gool}]{bossard14}
\bibinfo{author}{Bossard, L.}, \bibinfo{author}{Guillaumin, M.}, \bibinfo{author}{Van~Gool, L.}, \bibinfo{year}{2014}.
\newblock \bibinfo{title}{Food-101 -- mining discriminative components with random forests}, in: \bibinfo{booktitle}{European Conference on Computer Vision}.
\bibitem[{Bulatov(2011)}]{bulatov2011notmnist}
\bibinfo{author}{Bulatov, Y.}, \bibinfo{year}{2011}.
\newblock \bibinfo{title}{Notmnist dataset}.
\newblock \bibinfo{journal}{Google (Books/OCR), Tech. Rep.[Online]. Available: http://yaroslavvb. blogspot. it/2011/09/notmnist-dataset. html} \bibinfo{volume}{2}.
\bibitem[{Cimpoi et~al.(2014)Cimpoi, Maji, Kokkinos, Mohamed and Vedaldi}]{cimpoi2014describing}
\bibinfo{author}{Cimpoi, M.}, \bibinfo{author}{Maji, S.}, \bibinfo{author}{Kokkinos, I.}, \bibinfo{author}{Mohamed, S.}, \bibinfo{author}{Vedaldi, A.}, \bibinfo{year}{2014}.
\newblock \bibinfo{title}{Describing textures in the wild}, in: \bibinfo{booktitle}{Proceedings of the IEEE conference on computer vision and pattern recognition}, pp. \bibinfo{pages}{3606--3613}.
\bibitem[{Clanuwat et~al.(2018)Clanuwat, Bober-Irizar, Kitamoto, Lamb, Yamamoto and Ha}]{clanuwat2018deep}
\bibinfo{author}{Clanuwat, T.}, \bibinfo{author}{Bober-Irizar, M.}, \bibinfo{author}{Kitamoto, A.}, \bibinfo{author}{Lamb, A.}, \bibinfo{author}{Yamamoto, K.}, \bibinfo{author}{Ha, D.}, \bibinfo{year}{2018}.
\newblock \bibinfo{title}{Deep learning for classical japanese literature}.
\newblock \bibinfo{journal}{arXiv:1812.01718} .
\bibitem[{Cohen et~al.(2017)Cohen, Afshar, Tapson and Van~Schaik}]{cohen2017emnist}
\bibinfo{author}{Cohen, G.}, \bibinfo{author}{Afshar, S.}, \bibinfo{author}{Tapson, J.}, \bibinfo{author}{Van~Schaik, A.}, \bibinfo{year}{2017}.
\newblock \bibinfo{title}{Emnist: Extending mnist to handwritten letters}, in: \bibinfo{booktitle}{International Joint Conference on Neural Networks (IJCNN)}, \bibinfo{organization}{IEEE}. pp. \bibinfo{pages}{2921--2926}.
\bibitem[{D{\'\i}az-Rodr{\'\i}guez et~al.(2023)D{\'\i}az-Rodr{\'\i}guez, Del~Ser, Coeckelbergh, de~Prado, Herrera-Viedma and Herrera}]{diaz2023connecting}
\bibinfo{author}{D{\'\i}az-Rodr{\'\i}guez, N.}, \bibinfo{author}{Del~Ser, J.}, \bibinfo{author}{Coeckelbergh, M.}, \bibinfo{author}{de~Prado, M.L.}, \bibinfo{author}{Herrera-Viedma, E.}, \bibinfo{author}{Herrera, F.}, \bibinfo{year}{2023}.
\newblock \bibinfo{title}{Connecting the dots in trustworthy artificial intelligence: From ai principles, ethics, and key requirements to responsible ai systems and regulation}.
\newblock \bibinfo{journal}{Information Fusion} \bibinfo{volume}{99}, \bibinfo{pages}{101896}.
\bibitem[{Eshraghian et~al.(2023)Eshraghian, Ward, Neftci, Wang, Lenz, Dwivedi, Bennamoun, Jeong and Lu}]{eshraghian2021training}
\bibinfo{author}{Eshraghian, J.K.}, \bibinfo{author}{Ward, M.}, \bibinfo{author}{Neftci, E.}, \bibinfo{author}{Wang, X.}, \bibinfo{author}{Lenz, G.}, \bibinfo{author}{Dwivedi, G.}, \bibinfo{author}{Bennamoun, M.}, \bibinfo{author}{Jeong, D.S.}, \bibinfo{author}{Lu, W.D.}, \bibinfo{year}{2023}.
\newblock \bibinfo{title}{Training spiking neural networks using lessons from deep learning}.
\newblock \bibinfo{journal}{Proceedings of the IEEE} \bibinfo{volume}{111}, \bibinfo{pages}{1016--1054}.
\bibitem[{Fei-Fei et~al.(2006)Fei-Fei, Fergus and Perona}]{fei2006one}
\bibinfo{author}{Fei-Fei, L.}, \bibinfo{author}{Fergus, R.}, \bibinfo{author}{Perona, P.}, \bibinfo{year}{2006}.
\newblock \bibinfo{title}{One-shot learning of object categories}.
\newblock \bibinfo{journal}{IEEE transactions on pattern analysis and machine intelligence} \bibinfo{volume}{28}, \bibinfo{pages}{594--611}.
\bibitem[{Gerstner et~al.(2014)Gerstner, Kistler, Naud and Paninski}]{gerstner2014neuronal}
\bibinfo{author}{Gerstner, W.}, \bibinfo{author}{Kistler, W.M.}, \bibinfo{author}{Naud, R.}, \bibinfo{author}{Paninski, L.}, \bibinfo{year}{2014}.
\newblock \bibinfo{title}{Neuronal dynamics: From single neurons to networks and models of cognition}.
\newblock \bibinfo{publisher}{Cambridge University Press}.
\bibitem[{He et~al.(2016)He, Zhang, Ren and Sun}]{he2016deep}
\bibinfo{author}{He, K.}, \bibinfo{author}{Zhang, X.}, \bibinfo{author}{Ren, S.}, \bibinfo{author}{Sun, J.}, \bibinfo{year}{2016}.
\newblock \bibinfo{title}{Deep residual learning for image recognition}, in: \bibinfo{booktitle}{Proceedings of the IEEE conference on computer vision and pattern recognition}, pp. \bibinfo{pages}{770--778}.
\bibitem[{Helber et~al.(2019)Helber, Bischke, Dengel and Borth}]{helber2019eurosat}
\bibinfo{author}{Helber, P.}, \bibinfo{author}{Bischke, B.}, \bibinfo{author}{Dengel, A.}, \bibinfo{author}{Borth, D.}, \bibinfo{year}{2019}.
\newblock \bibinfo{title}{Eurosat: A novel dataset and deep learning benchmark for land use and land cover classification}.
\newblock \bibinfo{journal}{IEEE Journal of Selected Topics in Applied Earth Observations and Remote Sensing} .
\bibitem[{Hinton(2022)}]{hinton2022forward}
\bibinfo{author}{Hinton, G.}, \bibinfo{year}{2022}.
\newblock \bibinfo{title}{The forward-forward algorithm: Some preliminary investigations}.
\newblock \bibinfo{journal}{arXiv:2212.13345} .
\bibitem[{Kim and Panda(2021)}]{kim2021visual}
\bibinfo{author}{Kim, Y.}, \bibinfo{author}{Panda, P.}, \bibinfo{year}{2021}.
\newblock \bibinfo{title}{Visual explanations from spiking neural networks using inter-spike intervals}.
\newblock \bibinfo{journal}{Scientific Reports} \bibinfo{volume}{11}, \bibinfo{pages}{19037}.
\bibitem[{Krizhevsky et~al.(2009)Krizhevsky, Hinton et~al.}]{krizhevsky2009learning}
\bibinfo{author}{Krizhevsky, A.}, \bibinfo{author}{Hinton, G.}, et~al., \bibinfo{year}{2009}.
\newblock \bibinfo{title}{Learning multiple layers of features from tiny images} .
\bibitem[{Lake et~al.(2015)Lake, Salakhutdinov and Tenenbaum}]{lake2015human}
\bibinfo{author}{Lake, B.M.}, \bibinfo{author}{Salakhutdinov, R.}, \bibinfo{author}{Tenenbaum, J.B.}, \bibinfo{year}{2015}.
\newblock \bibinfo{title}{Human-level concept learning through probabilistic program induction}.
\newblock \bibinfo{journal}{Science} \bibinfo{volume}{350}, \bibinfo{pages}{1332--1338}.
\bibitem[{LeCun and Cortes(2010)}]{lecun_mnist_2010}
\bibinfo{author}{LeCun, Y.}, \bibinfo{author}{Cortes, C.}, \bibinfo{year}{2010}.
\newblock \bibinfo{title}{{MNIST} handwritten digit database} \URLprefix \url{http://yann.lecun.com/exdb/mnist/}.
\bibitem[{Lee and Song(2023)}]{lee2023symba}
\bibinfo{author}{Lee, H.C.}, \bibinfo{author}{Song, J.}, \bibinfo{year}{2023}.
\newblock \bibinfo{title}{Symba: Symmetric backpropagation-free contrastive learning with forward-forward algorithm for optimizing convergence}.
\newblock \bibinfo{journal}{arXiv:2303.08418} .
\bibitem[{Liang et~al.(2017)Liang, Li and Srikant}]{liang2017enhancing}
\bibinfo{author}{Liang, S.}, \bibinfo{author}{Li, Y.}, \bibinfo{author}{Srikant, R.}, \bibinfo{year}{2017}.
\newblock \bibinfo{title}{Enhancing the reliability of out-of-distribution image detection in neural networks}.
\newblock \bibinfo{journal}{arXiv:1706.02690} .
\bibitem[{Lillicrap et~al.(2016)Lillicrap, Cownden, Tweed and Akerman}]{lillicrap2016random}
\bibinfo{author}{Lillicrap, T.P.}, \bibinfo{author}{Cownden, D.}, \bibinfo{author}{Tweed, D.B.}, \bibinfo{author}{Akerman, C.J.}, \bibinfo{year}{2016}.
\newblock \bibinfo{title}{Random synaptic feedback weights support error backpropagation for deep learning}.
\newblock \bibinfo{journal}{Nature Communications} \bibinfo{volume}{7}, \bibinfo{pages}{13276}.
\bibitem[{Van~der Maaten and Hinton(2008)}]{van2008visualizing}
\bibinfo{author}{Van~der Maaten, L.}, \bibinfo{author}{Hinton, G.}, \bibinfo{year}{2008}.
\newblock \bibinfo{title}{Visualizing data using t-sne}.
\newblock \bibinfo{journal}{Journal of Machine Learning Research} \bibinfo{volume}{9}.
\bibitem[{Martinez-Seras et~al.(2023)Martinez-Seras, Del~Ser, Lobo, Garcia-Bringas and Kasabov}]{seras2022novel}
\bibinfo{author}{Martinez-Seras, A.}, \bibinfo{author}{Del~Ser, J.}, \bibinfo{author}{Lobo, J.L.}, \bibinfo{author}{Garcia-Bringas, P.}, \bibinfo{author}{Kasabov, N.}, \bibinfo{year}{2023}.
\newblock \bibinfo{title}{A novel out-of-distribution detection approach for spiking neural networks: Design, fusion, performance evaluation and explainability}.
\newblock \bibinfo{journal}{Information Fusion} \bibinfo{volume}{100}, \bibinfo{pages}{101943}.
\bibitem[{Merkel and Ororbia(2024)}]{merkel2024contrastive}
\bibinfo{author}{Merkel, C.}, \bibinfo{author}{Ororbia, A.G.}, \bibinfo{year}{2024}.
\newblock \bibinfo{title}{Contrastive learning in memristor-based neuromorphic systems}, in: \bibinfo{booktitle}{2024 IEEE Workshop on Signal Processing Systems (SiPS)}, \bibinfo{organization}{IEEE}. pp. \bibinfo{pages}{171--176}.
\bibitem[{Neftci et~al.(2019)Neftci, Mostafa and Zenke}]{neftci2019surrogate}
\bibinfo{author}{Neftci, E.O.}, \bibinfo{author}{Mostafa, H.}, \bibinfo{author}{Zenke, F.}, \bibinfo{year}{2019}.
\newblock \bibinfo{title}{Surrogate gradient learning in spiking neural networks: Bringing the power of gradient-based optimization to spiking neural networks}.
\newblock \bibinfo{journal}{IEEE Signal Processing Magazine} \bibinfo{volume}{36}, \bibinfo{pages}{51--63}.
\bibitem[{Netzer et~al.(2011)Netzer, Wang, Coates, Bissacco, Wu, Ng et~al.}]{netzer2011reading}
\bibinfo{author}{Netzer, Y.}, \bibinfo{author}{Wang, T.}, \bibinfo{author}{Coates, A.}, \bibinfo{author}{Bissacco, A.}, \bibinfo{author}{Wu, B.}, \bibinfo{author}{Ng, A.Y.}, et~al., \bibinfo{year}{2011}.
\newblock \bibinfo{title}{Reading digits in natural images with unsupervised feature learning}, in: \bibinfo{booktitle}{NIPS workshop on deep learning and unsupervised feature learning}, \bibinfo{organization}{Granada}. p.~\bibinfo{pages}{4}.
\bibitem[{Nilsback and Zisserman(2008)}]{nilsback2008automated}
\bibinfo{author}{Nilsback, M.E.}, \bibinfo{author}{Zisserman, A.}, \bibinfo{year}{2008}.
\newblock \bibinfo{title}{Automated flower classification over a large number of classes}, in: \bibinfo{booktitle}{2008 Sixth Indian conference on computer vision, graphics \& image processing}, \bibinfo{organization}{IEEE}. pp. \bibinfo{pages}{722--729}.
\bibitem[{Ororbia(2023)}]{ororbia2023learning}
\bibinfo{author}{Ororbia, A.}, \bibinfo{year}{2023}.
\newblock \bibinfo{title}{Contrastive-signal-dependent plasticity: Forward-forward learning of spiking neural systems}.
\newblock \bibinfo{journal}{arXiv preprint arXiv:2303.18187} .
\bibitem[{Ororbia and Mali(2023)}]{ororbia2023predictive}
\bibinfo{author}{Ororbia, A.}, \bibinfo{author}{Mali, A.}, \bibinfo{year}{2023}.
\newblock \bibinfo{title}{The predictive forward-forward algorithm}.
\newblock \bibinfo{journal}{arXiv:2301.01452} .
\bibitem[{Selvaraju et~al.(2017)Selvaraju, Cogswell, Das, Vedantam, Parikh and Batra}]{selvaraju2017grad}
\bibinfo{author}{Selvaraju, R.R.}, \bibinfo{author}{Cogswell, M.}, \bibinfo{author}{Das, A.}, \bibinfo{author}{Vedantam, R.}, \bibinfo{author}{Parikh, D.}, \bibinfo{author}{Batra, D.}, \bibinfo{year}{2017}.
\newblock \bibinfo{title}{Grad-cam: Visual explanations from deep networks via gradient-based localization}, in: \bibinfo{booktitle}{IEEE International Conference on Computer Vision (ICCV)}, pp. \bibinfo{pages}{618--626}.
\bibitem[{Stallkamp et~al.(2011)Stallkamp, Schlipsing, Salmen and Igel}]{stallkamp2011german}
\bibinfo{author}{Stallkamp, J.}, \bibinfo{author}{Schlipsing, M.}, \bibinfo{author}{Salmen, J.}, \bibinfo{author}{Igel, C.}, \bibinfo{year}{2011}.
\newblock \bibinfo{title}{The german traffic sign recognition benchmark: a multi-class classification competition}, in: \bibinfo{booktitle}{The 2011 international joint conference on neural networks}, \bibinfo{organization}{IEEE}. pp. \bibinfo{pages}{1453--1460}.
\bibitem[{Sun et~al.(2022)Sun, Ming, Zhu and Li}]{sun2022out}
\bibinfo{author}{Sun, Y.}, \bibinfo{author}{Ming, Y.}, \bibinfo{author}{Zhu, X.}, \bibinfo{author}{Li, Y.}, \bibinfo{year}{2022}.
\newblock \bibinfo{title}{Out-of-distribution detection with deep nearest neighbors}, in: \bibinfo{booktitle}{International Conference on Machine Learning}, \bibinfo{organization}{PMLR}. pp. \bibinfo{pages}{20827--20840}.
\bibitem[{Tavanaei et~al.(2019)Tavanaei, Ghodrati, Kheradpisheh, Masquelier and Maida}]{tavanaei2019deep}
\bibinfo{author}{Tavanaei, A.}, \bibinfo{author}{Ghodrati, M.}, \bibinfo{author}{Kheradpisheh, S.R.}, \bibinfo{author}{Masquelier, T.}, \bibinfo{author}{Maida, A.}, \bibinfo{year}{2019}.
\newblock \bibinfo{title}{Deep learning in spiking neural networks}.
\newblock \bibinfo{journal}{Neural networks} \bibinfo{volume}{111}, \bibinfo{pages}{47--63}.
\bibitem[{Terres-Escudero et~al.(2024)Terres-Escudero, Del~Ser and Garc{\'\i}a-Bringas}]{terres2024emerging}
\bibinfo{author}{Terres-Escudero, E.B.}, \bibinfo{author}{Del~Ser, J.}, \bibinfo{author}{Garc{\'\i}a-Bringas, P.}, \bibinfo{year}{2024}.
\newblock \bibinfo{title}{Emerging neohebbian dynamics in forward-forward learning: Implications for neuromorphic computing}.
\newblock \bibinfo{journal}{arXiv preprint arXiv:2406.16479} .
\bibitem[{Tosato et~al.(2023)Tosato, Basile, Ballarin, de~Alteriis, Cazzaniga and Ansuini}]{tosato2023emergent}
\bibinfo{author}{Tosato, N.}, \bibinfo{author}{Basile, L.}, \bibinfo{author}{Ballarin, E.}, \bibinfo{author}{de~Alteriis, G.}, \bibinfo{author}{Cazzaniga, A.}, \bibinfo{author}{Ansuini, A.}, \bibinfo{year}{2023}.
\newblock \bibinfo{title}{Emergent representations in networks trained with the forward-forward algorithm}.
\newblock \bibinfo{journal}{arXiv:2305.18353} .
\bibitem[{Xiao et~al.(2017)Xiao, Rasul and Vollgraf}]{xiao2017fashion}
\bibinfo{author}{Xiao, H.}, \bibinfo{author}{Rasul, K.}, \bibinfo{author}{Vollgraf, R.}, \bibinfo{year}{2017}.
\newblock \bibinfo{title}{Fashion-mnist: a novel image dataset for benchmarking machine learning algorithms}.
\newblock \bibinfo{journal}{arXiv:1708.07747} .
\bibitem[{Yamazaki et~al.(2022)Yamazaki, Vo-Ho, Bulsara and Le}]{yamazaki2022spiking}
\bibinfo{author}{Yamazaki, K.}, \bibinfo{author}{Vo-Ho, V.K.}, \bibinfo{author}{Bulsara, D.}, \bibinfo{author}{Le, N.}, \bibinfo{year}{2022}.
\newblock \bibinfo{title}{Spiking neural networks and their applications: A review}.
\newblock \bibinfo{journal}{Brain Sciences} \bibinfo{volume}{12}, \bibinfo{pages}{863}.
\bibitem[{Yang et~al.(2021)Yang, Zhou, Li and Liu}]{yang2021generalized}
\bibinfo{author}{Yang, J.}, \bibinfo{author}{Zhou, K.}, \bibinfo{author}{Li, Y.}, \bibinfo{author}{Liu, Z.}, \bibinfo{year}{2021}.
\newblock \bibinfo{title}{Generalized out-of-distribution detection: A survey}.
\newblock \bibinfo{journal}{arXiv:2110.11334} .
\bibitem[{Yang(2023)}]{yang2023theory}
\bibinfo{author}{Yang, Y.}, \bibinfo{year}{2023}.
\newblock \bibinfo{title}{A theory for the sparsity emerged in the forward forward algorithm}.
\newblock \bibinfo{journal}{arXiv preprint arXiv:2311.05667} .
\bibitem[{Zador et~al.(2022)Zador, Escola, Richards, {\"O}lveczky, Bengio, Boahen, Botvinick, Chklovskii, Churchland, Clopath et~al.}]{zador2022toward}
\bibinfo{author}{Zador, A.}, \bibinfo{author}{Escola, S.}, \bibinfo{author}{Richards, B.}, \bibinfo{author}{{\"O}lveczky, B.}, \bibinfo{author}{Bengio, Y.}, \bibinfo{author}{Boahen, K.}, \bibinfo{author}{Botvinick, M.}, \bibinfo{author}{Chklovskii, D.}, \bibinfo{author}{Churchland, A.}, \bibinfo{author}{Clopath, C.}, et~al., \bibinfo{year}{2022}.
\newblock \bibinfo{title}{Toward next-generation artificial intelligence: Catalyzing the neuroai revolution}.
\newblock \bibinfo{journal}{arXiv:2210.08340} .

\end{thebibliography}
\end{document}